\documentclass[12pt,reqno]{amsart}

\usepackage[utf8]{inputenc} % allow utf-8 input
\usepackage[T1]{fontenc}    % use 8-bit T1 fonts
\usepackage{url}            % simple URL typesetting
\usepackage{booktabs}       % professional-quality tables
\usepackage{amsfonts}       % blackboard math symbols
\usepackage{nicefrac}       % compact symbols for 1/2, etc.
\usepackage{microtype}      % microtypography

\usepackage{amsmath}
\usepackage{amsthm}
\usepackage{amssymb}
\usepackage[ruled]{algorithm2e}

\usepackage{stmaryrd}
\usepackage{mathrsfs}
\usepackage{graphicx}
\usepackage{mathtools}
\usepackage{wrapfig}
\usepackage{listings}
\usepackage{makecell}
\usepackage[shortlabels]{enumitem}

\usepackage[dvipsnames,table,xcdraw]{xcolor}
\usepackage[colorlinks=true,citecolor=JungleGreen,urlcolor=MidnightBlue,breaklinks=true]{hyperref}

\usepackage{natbib}
\usepackage[foot]{amsaddr}

\author{Armando J. {Cabrera~Pacheco}$^\ast$}
\email{a.cabrera@uni-tuebingen.de}
\thanks{$^\ast$ Equal contribution.}

\author{Rabanus Derr$^\ast$}
\email{rabanus.derr@uni-tuebingen.de}

\author{Robert C. Williamson} 
\address{University of T\"{u}bingen and T\"{u}bingen AI Center}
\email{bob.williamson@uni-tuebingen.de}
\newtheorem{theorem}{Theorem}[section]       % Fixed all these to use a single numbering scheme

\newtheorem{definition}[theorem]{Definition}
\newtheorem{corollary}[theorem]{Corollary}
\newtheorem{lemma}[theorem]{Lemma}
\newtheorem{remark}[theorem]{Remark}
\newtheorem{example}[theorem]{Example}

\usepackage{xspace}
\newcommand*{\eg}{e.g.\@\xspace}
\newcommand*{\ie}{i.e.\@\xspace}

\newcommand*{\cf}{cf.\@\xspace}
\newcommand*{\p}{p.\@\xspace}

\newcommand{\naturals}{\mathbb{N}}
\newcommand{\reals}{\mathbb{R}}

\newcommand{\aggregation}{\mathbf{A}}

\newcommand{\loss}{\lambda}

\def\pseudoprediction{pseudo-prediction }

\def\pseudopredictions{pseudo-predictions }
\def\Pseudopredictions{Pseudo-predictions }

\def\s{\sigma}
\def\w{\omega}

\def\g{\gamma}
\def\R{\mathbb{R}}
\def\To{\longrightarrow}
\def\learner{\textnormal{learner}}
\def\APA{\textnormal{APA}}
\def\defeq{\coloneqq}

\def\({\left(}
\def\){\right)}
\def\S{\Sigma}
\def\A{\mathbf{A}}
\def\Q{\mathbf{Q}}
\def\B{\mathbf{B}}
\def\tlambda{\widetilde{\lambda}}
\newcommand{\lm}[2]{\lim\limits_{#1\to #2}}
\DeclareMathOperator*{\argmin}{arg\,min}

\title[Axiomatic Loss Aggregation and an Adapted AA]{An Axiomatic Approach to Loss Aggregation \\ and an Adapted Aggregating Algorithm}

\begin{document}

\thispagestyle{empty}
\begin{abstract}
    Supervised learning has gone beyond the expected risk minimization framework. Central to most of these developments is the introduction of more general aggregation functions for losses incurred by the learner. In this paper, we turn towards online learning under expert advice. Via easily justified assumptions we characterize a set of reasonable loss aggregation functions as quasi-sums. Based upon this insight, we suggest a variant of the Aggregating Algorithm tailored to these more general aggregation functions. This variant inherits most of the nice theoretical properties of the AA, such as recovery of Bayes' updating and a time-independent bound on quasi-sum regret. Finally, we argue that generalized aggregations express the attitude of the learner towards losses.
\end{abstract}

\maketitle

{
\hypersetup{linkcolor=black}
\tableofcontents
}

\section{Introduction}\label{section-intro}
Whether it is framed as collaborative learning, distribution shift or data corruption, machine learning scholarship encountered the necessity to rethink the gold standard of learning theory: expected risk minimization.\footnote{This list is far from being exhaustive, \eg, fair machine learning \citep{williamson2019fairness}.} Central to this paradigm is the minimization of the average loss over a set of instances incurred by the learner. In the named more recent learning scenarios the average is replaced by other aggregation functionals which take into account the different data sources \citep[Eq. (1)]{haghtalab2022demand}, the distribution shift \citep[Eq. (R0)]{rahimian2019distributionally} or structured noise \citep{iacovissi2023general}. See \citep{frohlich2024risk} for a nice stratification of reasonable aggregations and their embedding in the literature on coherent risk measures and previsions. However, the generalization of the average aggregation remained within the borders of offline learning. An analogous development in online, adversarial learning does, to the best of our knowledge, not exist.

When focusing on learning under expert advice the main quantity of interest is the regret \citep{cesa2006prediction}. It compares the \emph{sum} of losses between learner and experts. In this work, we suggest a broader perspective on this sum of losses.

First, we put forward an axiomatical approach to reasonable loss aggregations in online learning, providing a generalized definition of regret. More concretely, under easily justified assumptions the aggregation forms a quasi-sum generated by an appropriate function $u \colon [0, \infty) \To [0, \infty)$, given by
\begin{align*}
    \Q^u_n(x_1, \ldots, x_n) =  u^{-1}\left(\sum_{i = 1}^n u(x_i)\right).
\end{align*}
Based upon this insight, we suggest a variant of the Aggregating Algorithm (AA) tailored to quasi-sums. The Aggregating Algorithm has first been suggested by \citet{vovk1990aggregating}. It  enjoys nice theoretical guarantees, \eg, a time-independent bound on regret and the recovery of Bayes' updating under an appropriate choice of loss function, while keeping simplicity. Its properties can, due to a ``change of variables'' in the loss, be transferred to our variant. Finally, we argue that generalized aggregations express the attitude of the learner towards losses incurred by providing predictions. In particular, we can tune the generator $u$ of the quasi-sum to express the aversity towards extreme losses (convex $u$) or the risk-seeking behavior of a learner (concave $u$). We provide experimental evidence corroborating this statement, which closes the loop by motivating the use of generalized aggregations in the first place.

% \end{itemize}
\section{A simple introduction to Aggregating Algorithm}
\label{an introduction to Aggregating Algorithm for dummies}

\begin{figure}
    \centering
    \def\svgwidth{0.8\columnwidth}
    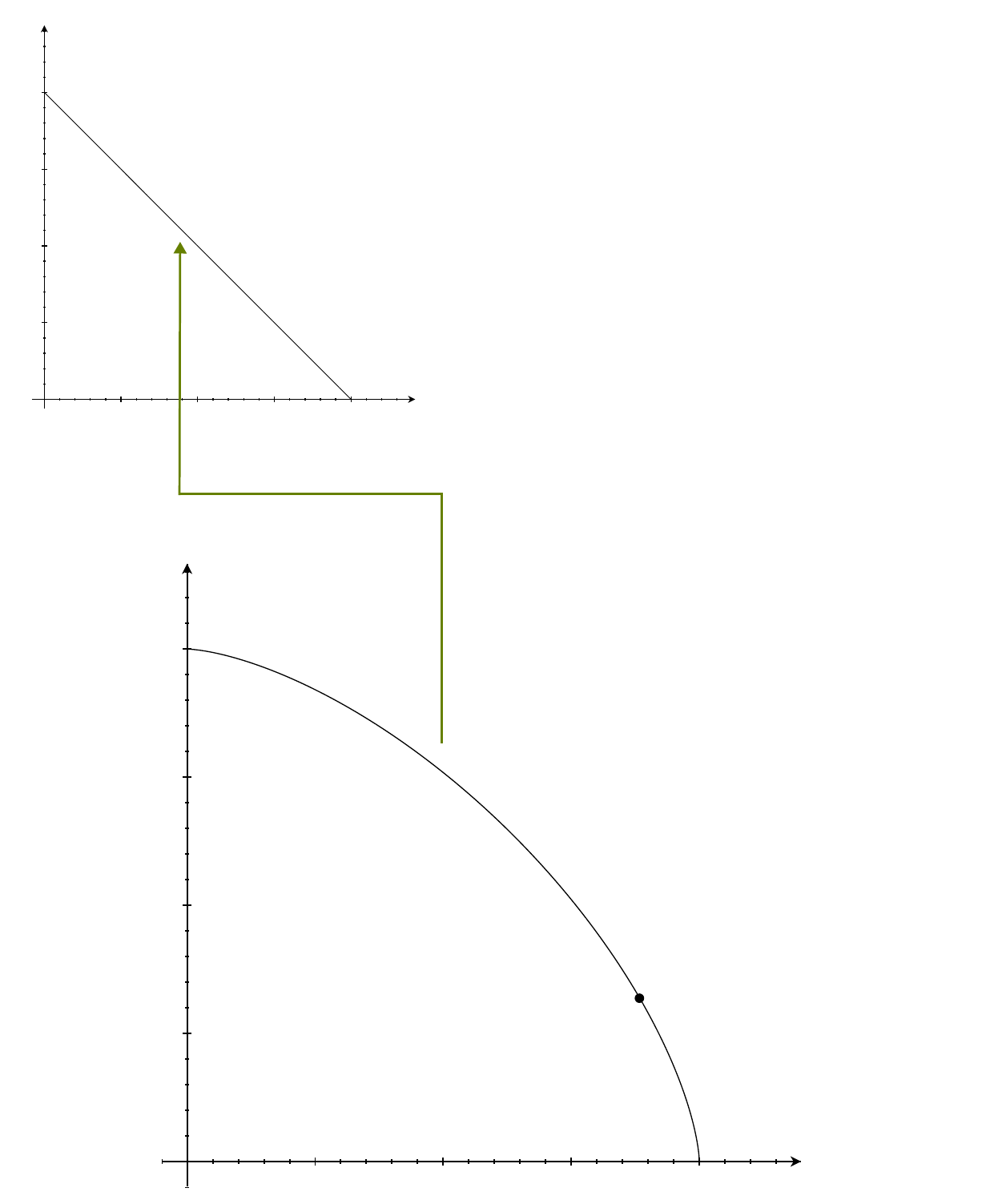
    \caption{Graphical Summary of the Steps in the Aggregating Algorithm. 
       {\tiny Experts $\theta_1$ and $\theta_2$ provide predictions $\xi(\theta_1)$ and $\xi(\theta_2)$, respectively, which are placed in the simplex (top-left) as $x_1 \coloneqq (\xi(\theta_1), 1- \xi(\theta_1))$ and $x_2 \coloneqq (\xi(\theta_2), 1- \xi(\theta_2))$ via $s \mapsto (s,1-s)$. The log-loss embeds the simplex as a curve in $\reals^2$ (top-right), \ie, $s \mapsto -\ln s$ is applied coordinate-wise and maps $x_1$ and $x_2$ to $x_1'$ and $x_2'$. Then, the exponential mapping projects them into $[0,1]^2$. The Aggregating Algorithm forms a convex combination $\psi$ of the projected predictions $x_1''$ and $ x_2''$ based on weights updated by a Bayesian-type formula (orange-brown), called a \pseudoprediction, which is substituted back to the simplex via a substitution function $\Sigma$ (darkgreen).}
    }
    \label{fig:aa intro}
\end{figure}

The Aggregating Algorithm is a relatively straightforward algorithm with strong theoretical guarantees. In order to motivate our theoretical development in this paper and to provide a low-threshold introduction to this algorithm, we first lead the reader through a sensibly simplified scenario of learning under expert advice.

Consider a binary class probability estimation task on the outcome space $\Omega \defeq \{ 0,1\}$. We repeatedly see $2$ experts $\Theta \defeq \{ \theta_1, \theta_2\}$ predicting the probability that the next outcome will be $1$, \ie, in every round $t \in [T] \coloneqq \{1, \ldots, T \}$ each expert $\theta \in \Theta$ provides a prediction $\xi_t(\theta) \in [0,1]$. After the predictions are given, the learner (who has seen the experts' predictions) has to commit to a prediction $\gamma_t \in [0,1]$ as well. Then, nature reveals an outcome $\w_t \in \{ 0, 1\}$. We measure the quality of the experts' and learner's prediction by the log-loss, that is, $\lambda(s, \w_t) = \llbracket \w_t = 0 \rrbracket \( -\ln(s) \) + \llbracket \w_t = 1 \rrbracket  \( -\ln(1- s) \)$, for $ s \in [0,1] $\footnote{We denote the Iveson-brackets as $\llbracket \ldots \rrbracket$.} ($s$~here refers to the prediction made either by the expert or the learner). Note that the set of possible predictions can be regarded as the probability simplex $s \mapsto (s ,1-s)$ for $s \in [0,1]$, cf. top-left in Figure~\ref{fig:aa intro}. Moreover, we can interpret the log-loss as an embedding of the simplex into $\mathbb{R}^2$, \ie, $(s, 1-s) \mapsto (-\ln(s), -\ln(1-s))$, see top-right in Figure~\ref{fig:aa intro}.

Now, the Aggregating Algorithm, as a \emph{learner}, uses the embedding of the simplex into $\mathbb{R}^2$ and \emph{exponentiates} it by the \emph{exponential mapping}  $(\lambda_0, \lambda_1) \mapsto (e^{-\eta \lambda_0 }, e^{-\eta \lambda_1})$, where $\eta>0$ is called the learning rate. Figure~\ref{fig:aa intro} right-top to bottom illustrates this step. In particular, the predictions of the experts can be traced through both mappings, the log-loss and the exponential mapping. The Aggregating Algorithm then forms a convex combination of the mapped experts' predictions $\psi_t$. The weights on how much each experts' prediction contributes to this mixture are based on a generalized Bayes' updating (see Section~\ref{Bayes' updating for weights}). The updating puts more weight on the experts which performed well in the past. As Figure~\ref{fig:aa intro} (orange-brown) illustrates, the obtained convex combination is not necessarily on the exponentiated embedding of the simplex anymore. We later call it a \emph{\pseudoprediction} for this reason. \Pseudopredictions have nice theoretical properties (see Section~\ref{A Generalized Aggregating Pseudo-Algorithm}), but they are not helpful as predictions, since they are not in the prediction space. That is why the Aggregating Algorithm requires a characteristical step: the substitution $\Sigma$ of the \pseudoprediction by an actual prediction with similar theoretical properties. This substitution $\Sigma$ is intuitively a projection of the \pseudoprediction $\psi$ to the exponentiated embedded simplex (\cf Figure~\ref{fig:aa intro} darkgreen). Crucially, a property called ``mixability'' (see Definition~\ref{def:u-mixability}) guarantees that the \pseudoprediction lies bottom-left to the exponentiated embedded simplex. The obtained actual predictions guarantee that the accumulated loss of the Aggregating Algorithm (learner), \ie, $\sum_{t = 1}^T \lambda(\gamma_t, \w_t)$ with $\gamma_t = \Sigma(\psi_t)$ is always smaller or equal than the accumulated loss of any expert, \ie, $\sum_{t = 1}^T \lambda(\xi_t(\theta), \w_t)$ for all $\theta \in \Theta$, up to a constant $C$. In other words, the Aggregating Algorithm's regret is bounded above by a constant, independent of the number of played rounds:

\begin{align*}
    \sum_{t = 1}^T \lambda(\gamma_t, \w_t) - \sum_{t = 1}^T \lambda(\xi_t(\theta), \w_t) \le C, \forall \theta \in \Theta.
\end{align*}

\section{Vovk's Aggregating Algorithm} \label{section-AA}
After this intuitive and illustrated introduction to Vovk's Aggregating Algorithm we now describe it in a rigorous way following \citep{vovk1990aggregating}. We also use this part to set up notation and remark its main properties. A \emph{prediction game} $(\Omega,\Gamma,\Theta,\lambda,\eta)$ consists of
\begin{description}
\item [Sample space] $\Omega$. This is regarded as the set of outcomes from nature. In general, we do not impose any structure on it. 
\item [Decision space] $\Gamma$. This is thought as the \emph{allowed predictions}. $\Gamma$ is a topological space endowed with the $\s$-algebra generated by the open sets.
\item [Parameter space] $\Theta$. We index the \emph{experts} (or \emph{decision strategies}) by $\theta \in \Theta$. We define $(\Theta, \mathcal{F}, \mu)$ as a measure space with $\mu$ being a base measure on the $\s$-algebra $\mathcal{F}$ on $\Theta$. For the sake of simplicity, we do not write out the base measure $\mu$ in the rest of the paper, \eg, we write the Lebesgue-integral for a measurable $E \in \mathcal{F}$,  $\mu(E) = \int_E d\theta$.
\item [Loss function] $\lambda\colon \Omega \times \Gamma \To [0,\infty]$. This gives us a way to measure the quality of the predictions.
\item [Learning rate] $\eta > 0$. A positive real number, typically as large as possible.
\end{description}
\begin{example}
\label{example:log-loss}
A widely used loss function in the prediction game is the log-loss $\lambda_{\ln} \colon \Omega \times \Gamma \To [0,\infty)$. In particular, when elements in $\Gamma$ are of the form $\g\colon \Omega \To [0,1]$, we define it as $\lambda_{\ln} (\w, \g) \coloneqq -\ln(\g(\w))$.
\end{example}

The prediction game works as follows. At each time $t \in [T]=\{1,...,T\} \subset \mathbb{N}$,
\begin{enumerate}
\item The experts make their predictions. That is, we have a measurable map $\xi_t \colon \Theta \To \Gamma$. $\xi_t(\theta)$ is interpreted by the prediction made by the expert $\theta$.
\item The learner  (who observes the predictions made by the experts) makes a prediction $\g_t \in \Gamma$.
\item Nature chooses the outcome $\w_t \in \Omega$.
\end{enumerate}

The goal of the learner is to ensure that its \emph{cumulative loss} is as good as the best expert's cumulative loss. In other words, we want to bound the \emph{regret},
\begin{align*}
    \textbf{R}_T(\theta) \defeq \mathbf{L}_T(\learner) - \mathbf{L}_T(\theta) \defeq \sum_{t=1}^T \lambda(\w_t,\g_t) -  \sum_{t=1}^T \lambda(\w_t,\xi_t(\theta)), \quad \forall \theta \in \Theta.
\end{align*}
Vovk's Aggregating Algorithm \citep{vovk1990aggregating, vovk2001competitive} (see Algorithm~\ref{alg:AA}) gives a bound which is independent of the number of played rounds $T$,
\begin{align} \label{eq-regretbound-AA-mix}
\mathbf{L}_T(\learner) -\mathbf{L}_T(\theta)  \leq  \frac{\ln n}{\eta}, \quad \forall \theta \in \Theta,
\end{align}
when the number of experts is finite ($n=|\Theta|<\infty$) and when the game is \emph{mixable}, \ie,  we assume the existence of a substitution function $\S$ for given $\lambda$ and $\eta$ such that
$%\label{eq-subs-function}
\lambda(\w,\S(\psi)) \leq \psi(\w),
$
for all $w \in \Omega$ and all \pseudopredictions $\psi$ of the form \eqref{eq-gen-pred}.\footnote{For several games, the log-loss (Example~\ref{example:log-loss}) can be shown to be mixable for $\eta \leq 1$.} As one can easily see mixability is required for the third step of the Algorithm~\ref{alg:AA}. Since the first two steps are somewhat independent of how the substitution concretely looks like they are sometimes summarized as \emph{Aggregating Pseudo-Algorithm (APA)}. The separation as well simplifies the analysis of the algorithm.
\begin{algorithm}
\label{alg:AA}
\caption{Aggregating Algorithm (AA)}
\KwData{Mixable prediction game $(\Omega,\Gamma,\Theta,\lambda,\eta)$ and prior distribution $P_0$.}
\KwResult{Predictions $(\gamma_t)$.}

\For{every $t$}{
    (1) Update experts' weights $P_t (\theta) \defeq e^{-\eta\lambda(\w_t,\xi_t(\theta))}P_{t-1}(\theta) $ \;
    (2) Provide \pseudoprediction,
    \makebox[\dimexpr \linewidth-4\algomargin][r]{% overlap left margin
        \begin{minipage}{\textwidth}% equation is in vmode
        \begin{align} \label{eq-gen-pred}
        \psi_t(\w) \defeq \ln_{e^{-\eta}} \left[    \int_{\Theta} e^{-\eta\lambda(\w_t,\xi_t(\theta))}P_{t-1}^*\,d\theta   \right], \text{ with }P_{t-1}^*(\theta) \defeq \frac{P_{t-1}(\theta)}{\int_{\Theta} P_{t-1}(\theta)\, d\theta} \text{.}
            \end{align}
        \end{minipage}\hspace{-\algomargin}}
    (3) Substitute the \pseudoprediction, $\gamma_t \coloneqq \S(\psi_t)$ such that $\lambda(\w,\S(\psi_t)) \leq \psi(\w)$, for all $\w \in \Omega$\;
}
\end{algorithm}

\subsection{Bayes' updating for weights} \label{Bayes' updating for weights}
A fundamental property of the AA is that, under some conditions, the updating scheme (Step (1) in AA (Algorithm~\ref{alg:AA})) is reduced to Bayesian updating \citep[Section 2.2]{vovk2001competitive}. More precisely, make the following choices: (a) $\Omega$ is finite, (b) $\Gamma = \Delta(\Omega)$, the set of all probability measures on $\Omega$, (c) loss function is the log-loss $\lambda_{\ln}$ and (d) the learning rate $\eta = 1$.

For an expert $\theta$ whose prediction at time $t$ is $\xi_t(\theta) \in \Delta(\Omega)$, we write $\xi_t(\theta)(\w_t)$ as the probability of seeing $\w_t$ forecasted by $\theta$, who has observed all data points up to time step $t-1$. Let $P_t (\theta)^*$ denote the normalized density $P_t (\theta)$:
\begin{align*}
    P_t (\theta)^* = \frac{e^{-\eta\lambda_{\ln}(\w_t,\xi_t(\theta))}P_{t-1}(\theta)}{\int e^{-\eta\lambda_{\ln}(\w_t,\xi_t(\theta))}P_{t-1}(\theta) d\theta}
    = \frac{\xi_t(\theta)(\w_t) P_{t-1}(\theta)}{\int \xi_t(\theta)(\w_t) P_{t-1}(\theta) d\theta}.
\end{align*}
One can interpret $\xi_t(\theta)(\w_t)$ as the likelihood to observe $\w_t$ at time $t$ given the expert $\theta$.
Analogously, $P_{t-1}(\theta)$ can be thought of as the prior on the experts after $t-1$ observations.
%The prediction $\xi_t(\theta)(\w_t)$ can be interpreted as the likelihood of observing $\w_t$ given expert $\theta$, who has observed all data points up to time step $t-1$. Analogously, $P_{t-1}(\theta)$ corresponds to the prior on the experts after $t-1$ observations.
Note that crucial in this derivation is the correspondence of the exponential projection $e^{-x}$ (with $\eta=1$) used in the definition of the \pseudopredictions and the log-loss $\lambda_{\ln}$.

We were led by preserving this reduction to Bayes' updating when adapting the AA to more general loss aggregations (\cf Section~\ref{A Generalized Aggregating Pseudo-Algorithm}).

\section{Generalizing the loss aggregation in learning under expert advice} \label{section-GenAgg}
The primary goal of the AA, as well as other algorithms solving the learning under expert advice problem, is to bound the summed loss of the learner by the summed loss of each of the experts plus an error term. We now turn towards an approach to loss aggregation from first principles. Instead of presuming the standard summation we provide a list of axioms for loss aggregation functionals which we readily justify (see for example  \citep{grabisch2009aggregation}).
\begin{definition}[Aggregation Functions]\label{def-agg-prop}
A function $\A \colon \bigcup_{n \in \naturals} [0,\infty)^n \To [0, \infty)$ is called an \emph{aggregation function}. We write $\A_n(x_1, \ldots, x_n)$ for the aggregation of $n$ instances. Let $x_1, \ldots, x_n, x \in [0, \infty)$. We define the following properties of $\A$.
\begin{enumerate}[label=\textnormal{(A\arabic*)}, ref=\textnormal{(A\arabic*)}]
    \item \label{thm:aggregation - continuity} {\bf Continuity.} We say $\A$ is \emph{continuous} if for every $x_i, i\in [n]$,
    \begin{align*}
    \lim_{x_i \To x}\A_n(x_1, \ldots, x_i, \ldots, x_n) = \A_n(x_1, \ldots, x, \ldots, x_n). 
    \end{align*}
    \item \label{thm:aggregation - strictly increasing} {\bf Monotonicity.} We say that $\A$ is \emph{strictly increasing} if for $x_i < x_i'$, $i\in [n]$, we have 
    \begin{align*}
    \A_n(x_1, \ldots, x_i, \ldots, x_n) < \A_n(x_1, \ldots, x_i', \ldots, x_n).
    \end{align*}
    \item \label{thm:aggregation - associative} {\bf Associativity.} We say that $\A$ is \emph{associative} if for all $x \in [0,\infty)$ and $i \in [n]$ we have 
    \begin{align*}
    \A_1(x) &= x,\\ 
    \A_n(x_1, \ldots, x_i, \ldots, x_n) &= \A_2(\A_{i}(x_1, \ldots, x_i), \A_{n-i}(x_{i+1}, \ldots, x_n)).
    \end{align*}  
    \item \label{thm:aggregation - loss compatibility} {\bf Loss compatibility}. We say that $\A$ is \emph{loss compatible} if $\A(0,...,0)=0$.
    % \item \label{thm:aggregation - Positive Homogeneity} {\bf Positive homogeneity.} We say that $\A$ is \emph{positively homogeneous} if for every $c \in [0,\infty)$ we have 
    % \begin{align*}
    % \A_n(c x_1, \ldots, c x_n) = c\A_n(x_1, \ldots, x_i, \ldots, x_n).
    % \end{align*}
\end{enumerate}
\end{definition}
Properties \ref{thm:aggregation - continuity}-\ref{thm:aggregation - loss compatibility} are natural to impose on an aggregation $\A$ of losses since they can be interpreted as follows. If $\A$ is continuous, an infinitesimally small change in a loss will result in an infinitesimal change in the aggregate of the losses. If it is strictly increasing, more loss on any instance give more aggregated loss. If it is associative, it is irrelevant how the losses are grouped together for aggregation. Note that even though this type of associativity immediately implies that the aggregation function is completely determined by the binary aggregation \citep[\p 33]{grabisch2009aggregation}, it does not directly imply commutativity.\footnote{For instance, consider the aggregation which gives back the last value, in terms of the intrinsic ordering of the input, of all elements.} Loss compatibility means that $0$ losses aggregate to $0$. It turns out we can fully characterize aggregation functions satisfying the properties above.

\begin{lemma}[Axiomatical Characterization of Loss-Aggregations]
\label{lemma-axiom-char}
    Let $\aggregation \colon \bigcup_{n \in \naturals} [0,\infty)^n \To [0, \infty)$ be an aggregation function. Suppose that $\A$ is continuous, strictly increasing, associative and loss compatible, \ie, it satisfies \ref{thm:aggregation - continuity} - \ref{thm:aggregation - loss compatibility}. Then, there exists a continuous, strictly increasing function $u \colon [0,\infty) \To [0, \infty)$, with $u(0) = 0$, such that
    \begin{align}\label{eq:axiom-characterization u-aggregation}
        \A_n(x_1, \ldots , x_n) = u^{-1}\left(\sum_{i = 1}^n u(x_i)\right).
    \end{align}
   If furthermore $\A$ is positively homogeneous, \ie, for every $c \in [0,\infty)$ and $n \in \naturals$ we have $\A_n(c x_1, \ldots, c x_n) = c\A_n(x_1, \ldots, x_i, \ldots, x_n)$, then $u(x) = x^k$ for some $k \in (0,\infty)$ in~\eqref{eq:axiom-characterization u-aggregation}.
   We call such aggregations \emph{quasi-sums} generated by $u$, and we denote them by $\Q^u$.
\end{lemma}
\begin{proof}
    Associativity~\ref{thm:aggregation - associative} guarantees that we can write
    \begin{align*}
        \A_n(x_1 , \ldots, x_n) = \A_2(x_1, \A_2(x_2, \A_2(\ldots))),
    \end{align*}
    for all $n \ge 2$ (\cf \citep[\p 33]{grabisch2009aggregation})

    In particular, 
    \begin{align*}
        \A_3(x_1, x_2, x_3) = \A_2(x_1, \A_2(x_2, x_3)) = \A_2(\A_2(x_1, x_2), x_3).
    \end{align*}
    Hence, $\A_2 \colon [0,\infty) \times [0, \infty) \To [0, \infty)$ is monotone, \ie, strictly increasing, continuous and associative in the sense of \citet{aczel1948operations} (for an English translation see \citep{aczel2012short}, for a different proof \citep{craigen1989associativity}), where it is shown that there exists $u \colon [0,\infty) \To [0, \infty)$ strictly increasing and continuous such that
    \begin{align*}
        \A_2(x_1, x_2) = u^{-1}\left( u(x_1) + u(x_2)\right).
    \end{align*}

    Since $0=\A_2(0,0) = u^{-1}(u(0)+u(0))$, it follows that $u(0)=0$.

   Finally, we obtain by induction (and associativity)
    \begin{align*}
        \A_n(x_1 , \ldots, x_n) &= \A_2(\A_{n-1}(x_1,...,x_{n-1}),x_n)    \\
        &= u^{-1}\( u\( u^{-1}\( \sum_{i=1}^{n-1} u(x_i)  \)\) + u(x_n)   \)  \\
        &=u^{-1}\( \sum_{i=1}^n u(x_i) \).
    \end{align*}

    For the second statement, we go back to $\aggregation^2$, which is now not only strictly increasing, continuous, associative and loss compatible, but as well positive homogeneous, \ie,
    \begin{align*}
        \A_2(c x_1, c x_2) = c \A_2(x_1, x_2),
    \end{align*}
    for all $c \in (0, \infty)$. Hence, Theorem 2 in \citep{aczel1955solution} (\cf \citep[\p 797]{gardner2018operations}) applies. This implies
    \begin{align*}
        \A_2(x_1, x_2) = \left( x_1^k + x_2^k \right)^{\frac{1}{k}},
    \end{align*}
    for some $k \in (0,\infty)$.
    By induction as above, we have
    \begin{align*}
        \A_n(x_1 , \ldots, x_n) &= \left( \sum_{i=1}^n x_i^k \right)^{\frac{1}{k}}.
        %&= \left( x_1^k + x_2^k + x_3^k + (\ldots)^k\right)^{\frac{1}{k}}.
    \end{align*}
\end{proof}
\begin{example}[$p$-Norms]
    Let $u(x) = x^p$ for $p>0$. Then $\Q_n^{u}(x_1, \ldots, x_n) = \left( \sum_{i = 1}^n x_i^p \right)^{1/p }$.
\end{example}

The statement here resembles a related older characterization of quasi-arithmetic means by \citet{kolmogorov1930notion} and \citet{nagumo1930klasse}.\footnote{Actually, quasi-sums and quasi-arithmetic means are related by performing the idempotization. See \citep[Section 6.5.1]{grabisch2009aggregation} for details.}

\begin{lemma}[Quasi-Sum to Quasi-Product]
\label{lemma:quasi-sum to quasi-product}
Let $\Q^u$ be a quasi-sum. Then, there exists a continuous, strictly decreasing function $f \colon [0,\infty) \To (0,1]$  with $f(0)=1$, such that 
\begin{align}
\Q_n(x_1,...,x_n) &=g\( f(x_1) \times ... \times f(x_n)   \), \label{eq-A-f-g}
\end{align}
where $g \colon (0,1]\To [0,\infty)$ is the inverse of $f$.\footnote{For the sake of readability, we neglect the multiplication sign in further expressions.}
\end{lemma}
\begin{proof}
Let $f(x) \defeq e^{-u(x)}$. It is straightforward to check that $f$ is strictly decreasing and that $f(0)=1$. Let $g(x)=u^{-1}(-\ln(x))$ be its inverse. Then, we can write
\begin{align*}
\Q_n(x_1,...,x_n) &= u^{-1}\( \sum_{i=1}^n u(x_i)  \) = g\(e^{- \( \sum_{i=1}^n u(x_i)  \)}    \)  =g\( f(x_1)...f(x_n)   \).
\end{align*}
\end{proof}

\begin{example}
Let $u(x)=x$, then $f(x)=e^{-x}$ and the corresponding aggregation \eqref{eq-A-f-g} is the sum:
\begin{align*}
\A_n(x_1,...,x_n) = g\( f(x_1) ... f(x_n)   \) = \mathbf{L}_n(x_1,...,x_n).
\end{align*}
\end{example}

In the remainder of this work we will sometimes write a quasi-sum $\Q^u$ as an aggregation function $\A$ of the form~\eqref{eq-A-f-g} when convenient. This will be explicitly stated or clear from context.

Equipped with this characterization of reasonable aggregation functionals of losses, it is natural to ask how can we solve the extended learning under expert advice problem: In which way do we have to modify existing online learning algorithms in order to provide regret guarantees such that the aggregated loss of the learner is bound above by the aggregated loss of any expert plus some error term?\footnote{The introduction of generalized aggregations to the literature on online learning, seems, to the best of our knowledge, a rather recent development. Closest to our work, we found \citep{neyman2023proper}. The authors use quasi-means, a related notion to our quasi-sums, to pool forecaster in a no-regret fashion. However, their loss aggregation is still the standard sum.} 

\section{The Aggregating Algorithm for quasi-sums}
In Section~\ref{section-GenAgg} we saw that a large class of reasonable aggregation functions can be expressed as quasi-sums. In this section, we show that one can still carry out the AA with respect to this type of aggregations (instead of the sum). To this end, we generalize the AA by preserving the reduction to Bayesian updating for an appropriate choice of loss and aggregation function.

We show the somewhat surprising property that this general AA leads to a ``change of variable" with respect to the loss function which can be interpreted as performing the standard AA with respect to a distorted loss, adding supporting evidence to the ``fundamentality of the log-loss'' (cf. \citep{vovk2015fundamental}). 

Although we want to derive a AA for quasi-sums, it turns out to be useful to see them as aggregations generated by a continuous, strictly decreasing function $f \colon [0,\infty) \To (0,1]$ with $f(0)=1$ (see \eqref{eq-A-f-g}). As it will be clear later, this function $f$ can be seen as the profile we choose to judge the experts.
\begin{definition}[Weighting Profile]
    \label{def:weighting profile}
    We say that a continuous $f \colon [0,\infty) \To (0,1]$ is a \emph{weighting profile} if (a) $f(0)=1$, (b) $f$ is strictly decreasing, and (c) $\lm{x}{\infty} f(x)=0$.
\end{definition}
It is straightforward to check that the the requirements we impose on the weighting profile are precisely the ones needed for the aggregation to be a quasi-sum. However, they can be justified independently. The normalization $f(0)=1$ means that weights should be positive and bounded from above by $1$. The expert incurring $0$ loss should get assigned full weight, while $f$ being strictly decreasing implies that the higher the loss the less weight should be put on the expert. The limiting behavior of $f$ says that an expert which incurs extremely large losses should be punished by getting down-weighted to $0$.

The properties of a weighting profile $f$ imply the existence of a continuous inverse, which we denote by $g \colon (0,1] \To [0,\infty)$, such that (a) $g(1)=0$, (b) $g$ is strictly decreasing, and (c) $\lm{x}{0^+} g(x)=\infty$.

For the rest of this work, for a fixed weighting profile $f$ (and hence its inverse $g$), we consider the aggregation function $\A \colon \bigcup_{n \in \naturals} [0,\infty)^n \To [0, \infty)$ given by
\begin{align} \label{eq-A}
\A_n(x_1,...,x_n) \defeq g(f(x_1)f(x_2)... f(x_n)).
\end{align}
Recall that using \eqref{eq-A-f-g} $\A$ can be expressed as a quasi-sum $\Q^u$ via the relation $f(x)=e^{-u(x)}$.

\subsection{An aggregating pseudo-algorithm for quasi-sums}
\label{A Generalized Aggregating Pseudo-Algorithm}
In order to adapt the AA to work with quasi-sums we first replace the Steps (1) and (2), \ie, the aggregating pseudo-algorithm, by an aggregating pseudo-algorithm for quasi-sums (APA-QS). We were led by preserving Bayesian updating when choosing $\lambda (x) = g(x)$ in Step (1).
\begin{algorithm}
\label{alg:APA-QS}
\caption{Aggregating Pseudo-Algorithm for Quasi-Sums (APA-QS)}
\KwData{Prediction game $(\Omega,\Gamma,\Theta,\lambda,\eta)$ and prior distribution $P_0$.}
\KwResult{Predictions $(\gamma_t)$.}

\For{every $t$}{
    (1) Update experts' weights $P_t (\theta) \defeq f(\lambda(\w_{t},\xi_{t}(\theta)))P_{t-1}(\theta)  \label{eq-updating-rule-GAA} $ \;
    (2) Provide \pseudoprediction with $P_{t-1}^*(\theta) \defeq \frac{P_{t-1}(\theta)}{\int_{\Theta} P_{t-1}(\theta)\, d\theta}$,
    \makebox[\dimexpr \linewidth-4\algomargin][r]{% overlap left margin
        \begin{minipage}{\textwidth}% equation is in vmode
        \begin{align}\label{eq-gen-prediction-APA-QS}
            \psi_t^f(\w)  = g\left[ \int_{\Theta} f(\lambda(\w,\xi_t(\theta))) P_{t-1}^*(\theta) \, d\theta    \right];
        \end{align}
        \end{minipage}\hspace{-\algomargin}}
    
}
\end{algorithm}

The \pseudoprediction allows for an interpretation as a normalizing factor for certain finite measure constructed using \eqref{eq-updating-rule-GAA}. More precisely, at each step $t$ we define the $\Omega$-dependent family of measures on $\Theta$, given by
\begin{align*}
p_t(\theta;\w) = f(\lambda(\w,\xi_t(\theta)))f(\psi_t^f(\w))^{-1}P_{t-1}^*(\theta),
\end{align*}
where $P_{t-1}^*$ is the normalization of $P_{t-1}$. Here, we do not specify $\psi_t \colon \Omega \To [0,\infty)$. 
Imposing $p_t(\theta;\w)$ to be a probability distribution on $\Theta$ yields to the \pseudoprediction given in \eqref{eq-gen-prediction-APA-QS}. Hence, we can interpret $\psi_t^f$ as the normalizing factor of $p_t(\theta;\w)$. For the sake of readability, we do not explicitly highlight the dependence of $\psi$ on $f$ in every instance. Instead, we use the $\psi^f$ notation in those cases where the dependency on $f$ is not clear or particularly important.
\begin{remark}
Suppose that $\w_t \in \Omega$ is revealed by nature. Then
\begin{align*}
\psi_t(\w_t)  = g\left[ \int_{\Theta} f(\lambda(\w_t,\xi_t(\theta))) P_{t-1}^*(\theta) \, d\theta    \right].
\end{align*}
If $\lambda(\w_t,\xi_t(\theta)) \gg 1$ for all $\theta \in \Theta$, by properties of the weighting profile $f$, the value of  $\psi_t(\w_t)$ will be very large. On the other hand, if $\lambda(\w_t,\xi_t(\theta)) \approx 0$, for all $\theta \in \Theta$, then its value will be close to 0. In this sense, we can interpret $\psi_t(\w_t)$ as the loss incurred by \pseudoprediction $\psi_t$ when $\w_t$ is observed (cf. \citep[Section 2.1]{vovk2001competitive}).
\end{remark}

\begin{lemma}[Bound for APA-Loss]
\label{lemma-APA}
Let $f \colon [0,\infty) \To \R$ be a weighting profile. Then
\begin{align*}
\A_T(\textnormal{APA}(P_0))  \defeq  \A_T\(\psi_1(\w_1),\psi_2(\w_2),...,\psi_T(\w_T) \) = g \left[ \int_{\Theta} f\(\A_T(\theta) \)  P_0(\theta)  \, d\theta  \right].
\end{align*}
Moreover, when $|\Theta|=n$ and $P_0$ is the uniform probability distribution with weights $1/n$, 
\begin{align}\label{eq-A-eta}
\A_T(\textnormal{APA}(P_0))  \leq \A_2 \( \A_T(\theta^*),g\(n^{-1}\) \),
\end{align}
for any expert $\theta^* \in \Theta$.
\end{lemma}
\begin{proof}
We will follow the general idea of the proof of Lemma 1 in \citep{vovk2001competitive}. Recall that using the updating rule \eqref{eq-updating-rule-GAA}, we have
\begin{align*}
P_{T} (\theta) &=f({\lambda(\w_{T},\xi_{T}(\theta))) ... f(\lambda(\w_{1},\xi_{1}(\theta))})P_0(\theta).
\end{align*}

It follows that:
\begin{align*}
f(\A_T(\theta)) P_0(\theta) &=f\(\lambda(\w_{T},\xi_{T}(\theta)) \) f\(\lambda(\w_{T-1},\xi_{T-1}(\theta))\) \hdots f\( \lambda(\w_{1},\xi_{1}(\theta))\) P_0(\theta) \\
&=f \(\lambda(\w_{T},\xi_{T}(\theta)) \) P_{T-1}(\theta) \frac{\int_{\Theta} P_{T-1}(\theta) \,d\theta}{\int_{\Theta} P_{T-1}(\theta) \,d\theta} \\
&=\int_{\Theta} P_{T-1}(\theta) \,d\theta \cdot f\( \lambda(\w_{T},\xi_{T}(\theta)) \) P_{T-1}^*(\theta)  \\
&=\int_{\Theta} P_{T-1}(\theta) \,d\theta \cdot  f(\psi_T(\w_T)) f\( \lambda(\w_{T},\xi_{T}(\theta)) \)f(\psi_T(\w_T))^{-1} P_{T-1}^*(\theta)  \\
&=\int_{\Theta} P_{T-1}(\theta) \,d\theta \cdot  f(\psi_T(\w_T)) p_T(\theta;\w_T).
\end{align*}

Integrating with respect to $\theta$ we obtain
\begin{align} \label{eq-bound-GAA-mixable}
\int_{\Theta} f\(\A_T(\theta) \)  P_0(\theta)  \, d\theta  = \int_{\Theta} P_{T-1}(\theta) \,d\theta \cdot  f(\psi_T(\w_T))
\end{align}

We now analyze $\int_{\Theta} P_{T-1}(\theta)\, d \theta$. Using similar arguments as above, we have
\begin{align*}
P_{T-1}(\theta) &= f(\lambda(\w_{T-1},\xi_{T-1}(\theta))) f(\lambda(\w_{T-2},\xi_{T-2}(\theta))) ... f(\lambda(\w_{1},\xi_{1}(\theta)))P_0(\theta) \\
&= f(\lambda(\w_{T-1},\xi_{T-1}(\theta))) P_{T-2}(\theta) \frac{\int_{\Theta} P_{T-2}(\theta) \,d\theta}{\int_{\Theta} P_{T-2}(\theta)\,d\theta} \\
&=\int_{\Theta} P_{T-2}(\theta) \,d\theta \cdot f(\psi_{T-1}(\w_{T-1})) f(\lambda(\w_{T-1},\xi_{T-1}(\theta)))f(\psi_{T-1}(\w_{T-1}))^{-1} P_{T-2}^*(\theta) \\
&=\int_{\Theta} P_{T-2}(\theta) \,d\theta \cdot f(\psi_{T-1}(\w_{T-1})) p_{T-1}(\theta; \w_{T-1}).
\end{align*} 

Integrating over $\Theta$ gives
\begin{align*}
\int_{\Theta} P_{T-1}(\theta) \, d\theta = \int_{\Theta} P_{T-2}(\theta) \,d\theta \cdot f(\psi_{T-1}(\w_{T-1})). 
\end{align*}

Continuing this process we arrive to

\begin{align} \label{eq-integral-equality}
\int_{\Theta} f\(\A_T(\theta) \)  P_0(\theta)  \, d\theta  &=\int_{\Theta} P_{0}(\theta) \,d\theta \cdot  f(\psi_1(\w_{1})) ... f(\psi_T(\w_{T})) \\
&=f(\psi_1(\w_{1})) ... f(\psi_T(\w_{T}))
\end{align}

Applying $g$ to both sides of \eqref{eq-integral-equality}, we obtain
\begin{align*}
g\left[ \int_{\Theta} f\(\A_T(\theta) \)  P_0(\theta)  \, d\theta  \right] = g\(  f(\psi_1(\w_1)) \hdots f(\psi_T(\w_T))   \) = \A_T(\textnormal{APA}(P_0)),
\end{align*}
as desired.

If $|\Theta|=n$ and $P_0$ is the uniform probability distribution with weights $1/n$ (cf. \cite{vovk1990aggregating}), we have for any fixed $\theta^* \in \Theta$,
\begin{align*}
g \left[ \int_{\Theta} f\(\A_T(\theta) \)  P_0(\theta)  \, d\theta  \right]  &=g \left[  \sum_{\theta=1}^n \frac{ f\(\A_T(\theta) \)}{n}  \right] \\
&\leq g \left[  \frac{ f\(\A_T(\theta^*) \)}{n}  \right] \\
&=g \left[   f\(\A_T(\theta^*) \) f\(g\(n^{-1}\)\)  \right]  \\
&=\A_2\(\A_T(\theta^*),g\(n^{-1}\)\).
\end{align*}
\end{proof}

Note that incorporating a learning rate $\eta >0$ in the APA (as in \citep{vovk2001competitive}) amounts to set $f(x)=e^{-\eta x}$.
\begin{corollary}\label{cor-GAA-learning rate}
    Let $f \colon [0,\infty) \To \R$ be a weighting profile. Let $\eta \in (0,\infty)$ be a learning rate %We define the inverse 
    and define $g_\eta(x) \coloneqq (f(x)^\eta)^{-1}$. When $|\Theta|=n$ and $P_0$ is the uniform probability distribution with weights $1/n$, 
    \begin{align}\label{eq-APA-bound}
    \A_T(\APA(P_0),\eta) =  \A_T(\textnormal{APA}(P_0))  \leq  \A_2\(\A_T(\theta^*),g_{\eta}\(n^{-1}\)\),
    \end{align}
    for any expert $\theta^* \in \Theta$.
\end{corollary}
\begin{proof}
Consider $f_{\eta}(x)=f(x)^{\eta}$ to define $\A^{\eta}$ (see \eqref{eq-A}), and notice that
\begin{align}\label{eq-A-power-inv}
\A^{\eta}(x_1,...x_n ) =g_{\eta}(f_{\eta}(x_1)...f_{\eta}(x_n)) 
=g(f(x_1)...f(x_n)) 
=\A(x_1,...,x_n).
\end{align}

Thus, applying Lemma~\ref{lemma-APA} with $\A^{\eta}$ we have
\begin{align*}
\A_T(\APA(P_0),\eta)\defeq\A_T^{\eta}(\textnormal{APA}(P_0))  = g_{\eta} \left[ \int_{\Theta} f_{\eta}\(\A_T^{\eta}(\theta) \)  P_0(\theta)  \, d\theta  \right].
\end{align*}

Further assuming that $|\Theta|=n$ and $P_0$ is the uniform probability distribution with weights $1/n$, we have (by the proof of Lemma~\ref{lemma-APA})

\begin{align*}
g_{\eta} \left[ \int_{\Theta} f_{\eta}\(\A_T^{\eta}(\theta) \)  P_0(\theta)  \, d\theta  \right] 
\leq \A_2^{\eta}\(\A_T^{\eta}(\theta^*),g_{\eta}\(n^{-1}\)\),
\end{align*}
for any $\theta^* \in \Theta$.

Using \eqref{eq-A-power-inv} again, we conclude that
\begin{align*}
\A_T(\APA(P_0),\eta)  \leq  \A_2\(\A_T(\theta^*),g_{\eta}\(n^{-1}\)\).
\end{align*}
\end{proof}
Notice that by setting $f(x)=e^{-x}$ in Corollary~\ref{cor-GAA-learning rate} we recover the bound found by \citet{vovk1990aggregating}.

\subsection{Using the APA-QS to make predictions}\label{section-APA-QS-pred}
As in the original AA \citep{vovk1990aggregating} (Step (3) in Algorithm~\ref{alg:AA}), in order to achieve the desired regret bound a necessary step is to turn the \pseudopredictions in to actual predictions (\ie, objects in $\Gamma$). This is achieved by using a substitution function. For this, fix $(\Omega,\Gamma,\theta,\lambda)$ and a weighting profile $f$. In this case, the \pseudopredictions are of the form
\begin{align} \label{eq-gen-prediction}
\psi(\w)  = g\left[ \int_{\Theta} f(\lambda(\w,\xi_t(\theta))) P_{t-1}^*(\theta) \, d\theta    \right]=g\left[ \int_{\Gamma} f(\lambda(\w,\g)) Q(\g) \, d\g  \right],
\end{align}
for some distribution $Q$ on $\Gamma$. We let $\mathcal{P}(\lambda,f)$ be the set of all \pseudopredictions of this form and define
\begin{align} \label{eq-c-f-dep}
c(f) \defeq \inf  \{ c \in \R \, | \, \forall \,  \psi \in \mathcal{P}(\lambda,f), \, \exists \, \g \in \Gamma, \, \forall \w, \, f(\lambda(\w,\g)) \geq  f(\psi(\w))^c   \},
\end{align}
and set $\inf\{\varnothing \} \defeq \infty$.

\begin{remark}
Note that if we fix $f$ and consider $f_{\eta}(x)=f(x)^{\eta}$, where $\eta>0$ is the learning rate, then we can consider the constant $c(f_{\eta})$ in \eqref{eq-gen-prediction} to depend only on $\eta$. When we do this we simply denote it by $c(\eta)$.
\end{remark}

When this infimum in \eqref{eq-c-f-dep} is attained, a substitution function $\S$ (which also depends on $\eta$ and $\lambda$) exists and satisfies
\begin{align} \label{eq-subs-rule-prop}
f(\lambda(\w,\S(\psi))) \geq f(\psi(\w))^{c(f)},
\end{align}
for all $\w \in \Omega$. For instance, \citet{vovk2001competitive} suggests the substitution function $\S(\psi) \in \argmin_{\gamma \in \Gamma} \sup_{\w \in \Omega} (u(\lambda(\w, \gamma)) / u(\psi(\w)))$.\footnote{When proper, mixable loss functions are considered in class probability estimation, \citet{williamson2023geometry}[Section 3.4] argue that antipolar losses constitute another universal substitution function.}

\begin{definition}[$(f, \eta)$-Mixability]
\label{def:u-mixability}
Let $(\Omega,\Gamma,\Theta,\lambda,\eta)$ be a prediction game. Let $f\colon [0, \infty) \To [0,1]$ be a weighting profile and consider \pseudopredictions $\psi \in \mathcal{P}(\lambda,f_{\eta})$ given by \eqref{eq-gen-prediction}. We call $(\Omega,\Gamma,\Theta,\lambda,\eta)$ \emph{$(f, \eta)$-mixable} if $c(\eta) = 1$, \ie, there exists a substitution function $\Sigma$ such that
\begin{align}\label{eq-mix-substitution}
    \lambda(\w, \Sigma(\psi)) \le \psi(\w),
\end{align}
for all $\w \in \Omega$. If the game is $(f, \eta)$-mixable for some $\eta$, we say the game is \emph{$f$-mixable}.
\end{definition}

Mixability guarantees that an actual prediction (that is, an element in $\Gamma$) can be found via a \pseudoprediction $\psi$. It can be shown that $c(f)\geq 1$ (Lemma~\ref{lemma-c-geq-1}). For the moment, we assume the game is mixable and derive some consequences and interpretations. First, note that in this case the learner using APA-QS (Algorithm~\ref{alg:APA-QS}) can apply the substitution function at each step $t \in [T]$ and hence ensures performing as good as the best expert. We call this procedure Aggregating Algorithm for quasi-sums (AA-QS). For the AA-QS we combine Definition~\ref{def:u-mixability} and Corollary~\ref{cor-GAA-learning rate} when $|\Theta|=n$, which implies the following result.
\begin{corollary}
Let $f \colon [0,\infty) \To \R$ be a weighting profile. Let $\eta \in (0,\infty)$ be a learning rate. When $(\Omega, \Gamma, \Theta, \lambda, \eta)$ is $(f,\eta)$-mixable, $|\Theta|=n$ and $P_0$ is the uniform probability distribution with weights $1/n$,
    \begin{align}\label{eq-f-GAPA-bound}
\A_T(\learner) &\defeq \A_T ( \lambda(\w_1, \Sigma(\psi_1)),...,\lambda(\w_T, \Sigma(\psi_T)))
%&\leq \A_T(\APA(P_0),\eta) \\
\leq  \A_2\(\A_T(\theta^*),g_{\eta}\(n^{-1}\)\),
\end{align}
for any expert $\theta^* \in \Theta$.
\end{corollary}
\begin{remark}
    The corollary requires the prediction game to be $(f, \eta)$-mixable. However, it turns out that we can generalize the bound on the aggregated loss of the learner to non-mixable losses (Theorem~\ref{thm-GAPA-bound}), which is analogous to \citep[Eq. (15)]{vovk1990aggregating}. Even the optimality result for the AA \citep{vovk1995game} is inherited by the AA-QS (Section~\ref{AA-optimality for quasi-sums}). Roughly summarized, it says that the constants in \eqref{eq-f-GAPA-bound} cannot be undercut by any other prediction algorithm.
\end{remark}
\begin{remark}
    The substitution function is a computational bottleneck which, in particular in high-dimensional prediction tasks, might lead to suboptimal time performance. We haven't found literature to address this shortcoming.
\end{remark}
\subsection{A change of variables for the AA} \label{sec-change-of-variables} Looking carefully at the bound \eqref{eq-f-GAPA-bound} reveals a somewhat surprising interpretation of performing the AA-QS with $\Q^u$. Let $(\Omega,\Gamma,\Theta,\lambda,\eta)$ be a prediction game with $|\Theta|=n$ and $c(\eta)=1$ (\ie, there exists a substitution rule satisfying \eqref{eq-mix-substitution}). Then, setting $f_{\eta}(x)=e^{-\eta u(x)}$ gives
\begin{align*}
\Q^{u}_T\(  \lambda(\w_1,\S(\psi_1))... \lambda(\w_T,\S(\psi_T)) \) \leq \Q^{u}_2 ( \Q^u_T(\theta^*),g_{\eta}(n^{-1}) ),
\end{align*}
where $g_{\eta}(x)=u^{-1}(-\ln(x)/\eta)$. Notice that after some immediate steps this is equivalent to
\begin{align*}
 \sum_{t=1}^T u\( \lambda(\w_t,\S(\psi_t))\)  \leq  \sum_{t=1}^T u\(\lambda(\w_t,\xi_t(\theta^*))\) + \frac{\ln\(n\)}{\eta},
\end{align*}
that is the bound provided by the original AA applied to the loss function $\tlambda \defeq u \circ \lambda$. Furthermore, we see right away that we can relate $(f,\eta)$-mixability with the usual $\eta$-mixability.
\begin{lemma}[$f$-Mixability is Mixability of Composite Loss]
\label{lemma:u-mixability is mixability of composite loss}
A loss $\lambda$ is $(f, \eta)$-mixable if and only if $\tlambda = u \circ \lambda$ for $u(x)\defeq -\ln f(x)$ is $\eta$-mixable.
\end{lemma}
\begin{proof}
    If $\lambda$ is $(f, \eta)$-mixable, there exists a substitution function $\Sigma$ such that for all $\psi \in \mathcal{P}(\lambda,f_{\eta})$,

    \begin{align*}
        \lambda(\w, \Sigma(\psi)) &\le \psi(\w), \forall \w \in \Omega\\
        \Longleftrightarrow \lambda(\w, \Sigma(\psi)) &\leq g_{\eta}\left[ \int_{\Gamma} f_{\eta}(\lambda(\w,\g)) Q(\g) \, d\g  \right], \forall \w \in \Omega\\
         \Longleftrightarrow \lambda(\w, \Sigma(\psi)) &\leq u^{-1} \(-\frac{\ln}{\eta}\left[ \int_{\Gamma} e^{-\eta u(\lambda(\w,\g))} Q(\g) \, d\g  \right]\), \forall \w \in \Omega\\
        \Longleftrightarrow u(\lambda(\w, \Sigma(\psi))) &\leq \ln_{e^{-\eta}}\left[ \int_{\Gamma} e^{-\eta u(\lambda(\w,\g))} Q(\g) \, d\g  \right], \forall \w \in \Omega
    \end{align*}
    Hence, $\Sigma$ is a substitution function for all $\psi \in \mathcal{P}(\tlambda, f_{\eta}(x)=e^{-\eta x})$, which is the standard $\eta$-mixability.
\end{proof}
Lemma~\ref{lemma:u-mixability is mixability of composite loss} implies the following corollary.
\begin{corollary}[Change of variables for the AA]
\label{corollary:change of variables of the AA}
Let $\mathcal{G} \defeq (\Omega,\Gamma,\Theta,\lambda,\eta)$ be a prediction game with $|\Theta|=n$ and $\lambda$ being $(f,\eta)$-mixable. Let $\widetilde{\mathcal{G}} \defeq (\Omega,\Gamma,\Theta,\tlambda,\eta)$ with $\tlambda = u \circ \lambda$ for $u(x)\defeq -\ln f(x)$. The AA-QS applied on $\mathcal{G}$ with aggregation $\A = \Q^u$ induced by $u$ achieves the same bound on the regret as the AA applied on $\widetilde{\mathcal{G}}$:
\begin{align}\label{eq-uAA-regret}
\Q^u_T(\learner)\defeq\Q^u_T( \lambda(\w_1,\S(\psi_1))... \lambda(\w_T,\S(\psi_T))  ) \leq u^{-1}\( u\(\Q^u_T(\theta)\) + \frac{\ln(n)}{\eta} \),
\end{align}
for any $\theta \in \Theta$.
\end{corollary}

\begin{remark}
It is worth to point out that usually one is interested in composing loss functions from the ``inside'', \ie, reparametrizations. Here the transformation is extrinsic, we push the loss curve via $u$ obtaining a ``distorted'' version of the original loss function $\lambda$.
\end{remark}

\begin{remark}
    \citet{vovk2015fundamental} argues that the log-loss is in a particular way \emph{fundamental}. As \citet{pacheco2023geometry}  have shown, both Vovk's fundamentality and mixability (in the sufficiently differentiable case) are equivalent to a curvature comparison between a given loss and the log-loss. Definition~\ref{def:u-mixability} and the analysis of the AA with more general aggregation functions, emphasize that the choice of aggregation is tightly intertwined with the definition of mixability. Particularly, standard sum aggregation corresponds to standard mixability. Hence, the fundamentality of the log-loss is a particularity of the standard sum aggregation and does not imply that the log-loss is fundamental for other aggregations. However, Corollary~\ref{corollary:change of variables of the AA} suggest a connection back to a type fundamentality of log-loss via $u$, which can potentially be described geometrically. However, as discussed in Section~\ref{sec-change-of-variables}, whether a constant regret bound for the aggregation $\Q^u$ holds also  depends on whether $\tlambda = u \circ \lambda$ is mixable (see Section~\ref{section-AA}) hinting to another way of considering the log-loss \emph{fundamental}. We will not go deeper into this observation here.%\arm{I need to check this}
\end{remark}

\section{AA-optimality for quasi-sums}
\label{AA-optimality for quasi-sums}
So far, we have assumed that the prediction game in consideration is $(f,\eta)$-mixable. In this case we obtain a direct bound for the aggregated loss of the predictions given via the substitution function and the APA-QS. As it happens with the usual AA, there is no guarantee that the given loss will be $(f,\eta)$-mixable. We deal with this situation in this section. With this at hand we also provide an optimality result for the Aggregating Algorithm under quasi-sum aggregation.

First, it will be useful to impose some mild conditions on the prediction game.
\begin{definition}[Regular Local Prediction Game \citep{vovk1995game}]\label{def-local-game}
    We call the tuple $(\Omega, \Gamma, \lambda)$ a \emph{local prediction game}. A local prediction game is called \emph{regular} if the following four assumptions hold
    \begin{enumerate}[(a)]
        \item $\Gamma$ is a compact topological space.
        \item For each $\omega \in \Omega$ the function $\gamma \mapsto \lambda(\omega, \gamma)$ is continuous.
        \item There exists $\gamma \in \Gamma$ such that, $\lambda(\omega, \gamma) < \infty$ for all $\omega \in \Omega$.
        \item For all $\gamma \in \Gamma$ there exists $\omega \in \Omega$ such that $\lambda(\omega, \gamma) \neq 0$.
    \end{enumerate}
\end{definition}
In this section it will be always assumed that the $(\Omega, \Gamma, \lambda)$ is a local prediction game.

\subsection{Aggregation algorithm for non-mixable losses}

Fix $(\Omega,\Gamma,\Theta,\lambda)$ and a weighting profile $f$, as in Section~\ref{section-APA-QS-pred}. Note, we assume that we obtain a regular local prediction game when we drop the set of experts $\Theta$. Recall that in this case, the \pseudopredictions belong to $\mathcal{P}(\lambda,f)$, that is, they are of the form
\begin{align*} 
\psi(\w)  =g\left[ \int_{\Gamma} f(\lambda(\w,\g) Q(\g) \, d\g  \right],
\end{align*}
for some distribution $Q$ on $\Gamma$.

First we show that $c \geq 1$ as in \citep{vovk1990aggregating}. The proof is basically the same and we include it for the reader's convenience. 
\begin{lemma}\label{lemma-c-geq-1}
For given $(\Omega,\Gamma,\theta,\lambda)$ and a weighting profile $f$, then $c(f)\geq 1$.
\end{lemma}
\begin{proof}
Suppose that there is $f$ such that $c \defeq c(f) < 1$. Let $\g' \in \Gamma$ and $Q_{\g'}$ be defined as $Q_{\g'}(\g')=1$ and $0$ otherwise. Then, there exists $\g \in \Gamma$ such that for all $\w \in \Omega$,
\begin{align*}
f(\lambda(\w,\g)) &\geq f\left( g \left[ \int_{\Gamma} f(\lambda(\w,\g)) \, Q_{\g'} \, d\Gamma    \right]   \right)^c \\
&\geq f(\lambda(\w,\g '))^c.
\end{align*}

Since $0 < f(x) \leq 1$ for all $x$, we have
\begin{align*}
0 \leq -\ln\(f(\lambda(\w,\g))  \) \leq c \( -\ln\(f(\lambda(\w,\g '))\) \) .
\end{align*}

By assumption (c), we know there exists $\g_1 \in \Gamma$ such that $\lambda(\w,\g_1) < \infty$. By the argument above (with $\g'=\g_1$) we know there is $\g_2 \in \Gamma$ such that
\begin{align*}
0 \leq -\ln\( f(\lambda(\w,\g_2))  \) \leq c \( -\ln \( f(\lambda(\w,\g_1)) \)\).
\end{align*} 

Continuing this way, we obtain a sequence $\{\g_k\} \subset \Gamma$ such that

\begin{align*}
0 \leq -\ln\(f(\lambda(\w,\g_{k+1}))  \) \leq c \( -\ln\(f(\lambda(\w,\g_k))\)\).
\end{align*}

Using the compactness of $\Gamma$ (assumption (a)), let $\g \in \Gamma$ be a limit point of this sequence. By continuity (assumption (b)), $-\ln \( f(\lambda(\w,\g))\) $ is the limit of a subsequence $\{ -\ln\(f(\lambda(\w,\g_{k})) \) \}$. 

Note that we have
\begin{align*}
0 \leq  -\ln\(f(\lambda(\w,\g_{k})) \) \leq c^{k-1}\( -\ln\(f(\lambda(\w,\g_1)) \) \),
\end{align*}
thus when $k \to \infty$ we conclude that $ \ln\(f(\lambda(\w,\g)\)=0$, that is $\lambda(\w,\g)=0$, which contradicts~(d).
\end{proof}

We are now ready to obtain an analogous bound to \eqref{eq-f-GAPA-bound}.
\begin{theorem}\label{thm-GAPA-bound}
For given $(\Omega,\Gamma,\theta,\lambda)$ and a weighting profile $f$. Let $c\defeq c(f)$. When $|\Theta|=n$ and $P_0$ is the uniform probability distribution with weights $1/n$, we have the bound
\begin{align}\label{eq-bound-GAPA}
 \A_T(\learner) \le \A_T(\textnormal{APA-QS}(P_0))  \leq  g\left[  \frac{f(\A_T(\theta^*))^{c}}{n^c} \right].
\end{align}
for any expert $\theta^* \in \Theta$. 

Moreover, if we consider a learning rate $\eta > 0$, $f_{\eta}(x)=f(x)^{\eta}$ and $c_{\eta} \defeq c(f_{\eta})$, we have
\begin{align}\label{eq-bound-GAPA-eta}
 \A_T(\learner) \le \A_T(\textnormal{APA}(P_0))  \leq g(f(\A_T(\theta^*))^{c_{\eta}} f(g_\eta(n^{-1}))^{c_{\eta}}),
\end{align}
where $g_{\eta}$ is the inverse of $f_{\eta}$.
\end{theorem}

\begin{proof}
Let $\psi$ be a \pseudoprediction. Then, we have $f(\lambda(\w,\g) \geq f(\psi(\w))^c$.
\begin{align*}
 \A_T(\learner) = g(f(\lambda(\w_1,\g_1)...f(\lambda(\w_T,\g_T)) \leq g(f(\psi_1(\w_1))^c...f(\psi_T(\w_T))^c).
\end{align*}
%\arm{perhaps one can avoid using $\B$ but might be a pain to rewrite the part below}
This motivate us to define a new aggregation function given by
\begin{align*}
\B^{c}_n(x_1,...,x_n) \defeq g\(f(x_1)^{c}...f(x_n)^{c}\).
\end{align*}

Using the fact that $c \geq 1$ (Lemma~\ref{lemma-c-geq-1}) and the notation in Lemma~\ref{lemma-APA}, we see that

\begin{align*}
f(\B^c(\theta)) P_0(\theta) &= f(\lambda(\w_1,\xi_1(\theta))^c....f(\lambda(\w_T,\xi_T(\theta))^c P_0( \theta ) \\
&\leq f(\lambda(\w_1,\xi_1(\theta))....f(\lambda(\w_T,\xi_T(\theta)) P_0( \theta ) 
\end{align*}

Proceeding as the in the proof of Lemma~\ref{lemma-APA}, we obtain

\begin{align*} 
\int_{\Theta} f\(\B_T^c(\theta) \)  P_0(\theta)  \, d\theta & \leq f(\psi_1(\w_{1})) ... f(\psi_T(\w_{T})),
\end{align*}

and hence,
\begin{align}\label{eq-int-bound-GAPA}
g\left[\int_{\Theta} f\(\B_T^c(\theta) \)  P_0(\theta)  \, d\theta \right] &  \geq g\(f(\psi_1(\w_{1})) ... f(\psi_T(\w_{T}))\)= \A_T(\textnormal{APA}(P_0)).
\end{align}

We are left to estimate the LHS of \eqref{eq-int-bound-GAPA}:

\begin{align*}
g\left[ \int_{\Theta} f\(\B_T^{c}(\theta) \)  P_0(\theta)  \, d\theta  \right] &= g\left[ \int_{\Theta} f\(\A_T(\theta) \)^{c}  P_0(\theta)  \, d\theta  \right]   \\
\leq& g\left[  \frac{f(\A_T(\theta^*))^{c}}{n} \right] \\
\leq & g\left[  \frac{f(\A_T(\theta^*))^{c}}{n^c} \right],
\end{align*}
proving \eqref{eq-bound-GAPA}.

To obtain \eqref{eq-bound-GAPA-eta}, let $f=f_{\eta}$ and $c_{\eta} \defeq c(f_{\eta})$, then we have
\begin{align*}
g_{\eta}\left[ \int_{\Theta} f_{\eta}\(({\B^{\eta}})_T^{c_{\eta}}(\theta) \)  P_0(\theta)  \, d\theta  \right] &= g_{\eta}\left[ \int_{\Theta} f_{\eta}\(\A^{\eta}_T(\theta) \)^{c_{\eta}}  P_0(\theta)  \, d\theta  \right] \leq g_{\eta} \left[  \frac{f_{\eta}(\A^{\eta}_T(\theta^*))^{c_{\eta}}}{n^{c_{\eta}}} \right],
\end{align*}
where $(\B^{\eta})_n^{c}(x_1,...,x_n) \defeq g_{\eta} (f_{\eta}(x_1)^c,...,f_{\eta}(x_n)= \B_n^c(x_1,...,x_n)$. 

The result follows since
\begin{align*}
g_{\eta} \left[  \frac{f_{\eta}(\A_T^{\eta}(\theta^*))^{c_{\eta}}}{n^{c_{\eta}}} \right] &= ({\B^{\eta}})_2^{c_{\eta}} ( \A^{\eta}_T(\theta^*),g_\eta(n^{-1}) )\\
&=  \B_2^{c_{\eta}} ( \A_T(\theta^*),g_\eta(n^{-1}) ) \\
=& g(f(\A_T(\theta^*))^{c_{\eta}} f(g_\eta(n^{-1}))^{c_{\eta}}).
\end{align*}
\end{proof}

\begin{remark}\label{remark-non-mixable-aa-with-u}
Recall that the weighting profile $f$ in Theorem~\ref{thm-GAPA-bound} can be written in the form $f(x)=e^{-u(x)}$ for an appropriate $u$. In this case, for $\eta>0$, we have
\begin{align*}
 g(f(\A_T(\theta^*))^{c_{\eta}} f(g_\eta(n^{-1}))^{c_{\eta}}) &= u^{-1}\(c_{\eta} u(\A_T(\theta^*)) +  c_{\eta} u(g_\eta(n^{-1}))\) \\
&= u^{-1}\(c_{\eta} u(\A_T(\theta^*)) +  \frac{c_{\eta}}{\eta} \ln(n)\).
\end{align*}

In particular, when $u(x)=x$, this gives
\begin{align*}
\mathbf{L}_T(\APA(P_0,\eta)) \leq c_{\eta} \mathbf{L}_T(\theta^*) + c_{\eta} \frac{\ln(n)}{\eta},
\end{align*}
as in \citep{vovk1990aggregating}.
\end{remark}

\subsection{AA-optimality}
Surprisingly, it is possible to show that the Aggregating Algorithm is in a game-theoretic sense optimal \citep{vovk1995game}. In a general game between an adversarial environment, which gets to choose experts' predictions and nature's outcome and a learner, the learner can only win if they achieve the bounds which are suggested by the Aggregating Algorithm, cf. Remark~\ref{remark-non-mixable-aa-with-u}. We formalize this statement and extend it to more general aggregation functions in the following.
\begin{definition}\label{def-global-pred-game}
    Let $\aggregation$ be a continuous, strictly increasing and associative aggregation function. Let $(\Omega, \Gamma, \lambda)$ be a regular local prediction game.
    We call the following full-information game $\mathfrak{G}$ between environment $E$ and learner $L$ a \emph{global prediction game}:
    \begin{enumerate}
        \item $E$ chooses the size $n$ of a set of experts $\Theta$.
        \item For every $t \in [T]$,
        \begin{enumerate}[(i)]
            \item $E$ chooses predictions $\xi_t(\theta) \in \Gamma$ for every $\theta \in \Theta$.
            \item $L$ chooses a prediction $\gamma_t \in \Gamma$.
            \item $E$ chooses an outcome $\omega_t \in \Omega$.
            \item  $\aggregation_t(\theta) \coloneqq \aggregation_{2}(\aggregation_{t-1}(\theta), \lambda(\omega_t, \xi_t(\theta)))$ for all $\theta \in \Theta$.\footnote{we set $\A_0(\theta) = 0$ for all $\theta \in \Theta$.}
            \item $\aggregation_t(\learner) \coloneqq \aggregation_{2}(\aggregation_{t-1}(\learner), \lambda(\omega_t, \gamma_t))$.
        \end{enumerate}
    \end{enumerate}
\end{definition}
\begin{definition}\label{def-AA-opt-bound}
We say that the learner $L$ \emph{wins the global prediction game} $\mathfrak{G}$ if for all $t \in [T]$ and $\theta \in \Theta$ there are constants $c$ and $a$ such that
\begin{align} \label{eq-regretbound-GAA}
\aggregation_t(\learner) \leq u^{-1}( c \ u(\A_t(\theta)) + a  \ln(n) ),
\end{align}
otherwise, we say that \emph{nature wins}. Note, that the aggregation function is a quasi-sum with generator $u$, \ie, $\aggregation = \Q_u$.
\end{definition}
Optimality here is grounded in the global game specified above. Intuitively, the following theorem shows that in a worst-case scenario, concerning the choice of experts and outcomes, every learner under expert advice can at best achieve the regret bound parametrized by $c$ and $a$ of \eqref{eq-regretbound-GAA}. Note that we don't put any restrictions on the abilities of the learner until this point. Strikingly, the Aggregating Algorithm can achieve this regret bound, hence is optimal.

\begin{theorem}[Optimality of Constant Regret Bound for All Predictors]
    Let $\aggregation$ be a continuous, strictly increasing and associative aggregation function. Consider the global prediction game following Definition~\ref{def-global-pred-game}. There exists a learner $L$ which against an arbitrary adversarial environment wins, \ie, for all $T \in \naturals$ and all $\theta \in \Theta$,
    \begin{align*} 
        \aggregation_T(L) \coloneqq \aggregation_T(\lambda(\omega_1, \gamma_1),\ldots, \lambda(\omega_T, \gamma_T)) \le u^{-1}( c \ u(\A_T(\theta^*)) + a  \ln(n) ),
    \end{align*}
    if and only if $c \ge c(\eta)$ and $a \ge \frac{c(\eta)}{\eta}$ for some $\eta \in [0,\infty)$ with $c(\eta)$ as defined in \eqref{eq-c-f-dep}, and $u$ is the generator of the aggregation, \ie, $\aggregation = \Q_u$.
\end{theorem}
\begin{proof}
First, we note that the aggregation $\aggregation$ fulfills \ref{thm:aggregation - continuity}-\ref{thm:aggregation - associative}. Hence, $\aggregation = \Q^u$ for some generator $u\colon [0,\infty) \To [0,\infty)$, continuous and strictly increasing with $u(0) = 0$ (Lemma~\ref{lemma-axiom-char}).

Let us define the surrogate loss $\tlambda \coloneqq u \circ \lambda$. It is straightforward to check that $(\Omega, \Gamma, \tlambda)$ fulfills all conditions for a regular local prediction game. The first condition holds by assumption. The second condition is clear, since the composition of continuous functions is continuous. Thirdly, there exists $\gamma \in \Gamma$ such that, $\lambda(\omega, \gamma) < \infty$ for all $\omega \in \Omega$. It follows $\tlambda(\omega, \gamma) = u(\lambda(\omega, \gamma)) < \infty$ for all $\omega \in \Omega$. Finally, for all $\gamma \in \Gamma$ there exists $\omega \in \Omega$ such that $\lambda(\omega, \gamma) \neq 0$, hence for all $\gamma \in \Gamma$ there exists $\omega \in \Omega$ such that $\tlambda(\omega, \gamma) = u(\lambda(\omega, \gamma)) \neq 0$, because $u(0) = 0$ and $u$ strictly increasing. Concluding, $(\Omega, \Gamma, \tlambda)$ is a regular local prediction game.

Theorem 1 in \citep{vovk1995game} states that in the specified game the learner $L$ is guaranteed to achieve the regret bound, for all $T \in \naturals$ and all $\theta \in \Theta$,
\begin{align}
\label{eq-best-regret-bound-possible}
 \sum_{t=1}^T \tlambda(\w_t,\gamma_t)) \le c \sum_{t=1}^T \tlambda(\w_t,\xi_t(\theta))) + a\ln\(|\Theta|\),
\end{align}
if and only if $c \ge \widetilde{c}(\eta)$ and $a \ge \frac{\widetilde{c}(\eta)}{\eta}$ for some $\eta \in [0,\infty)$, where
\begin{align*}
\widetilde{c}(\eta) \defeq \inf  \{ c \in \R \, | \, \forall \,  \psi \in \mathcal{P}(\Tilde{\lambda},e^{-\eta x}), \, \exists \, \g \in \Gamma, \, \forall \w, \, e^{-\eta\Tilde{\lambda}(\w,\g)} \geq  e^{{-\eta\psi(\w)}^c}  \},
\end{align*}
and set $\inf\{\varnothing \} \defeq \infty$, \cf \eqref{eq-c-f-dep}.
Note, that
\begin{align*}
    \widetilde{c}(\eta) &= \inf  \{ c \in \R \, | \, \forall \,  \psi \in \mathcal{P}(\lambda,e^{-\eta u(x)}), \, \exists \, \g \in \Gamma, \, \forall \w, \, e^{-\eta u(\lambda(\w,\g))}\geq  e^{{-\eta u(\psi(\w))}^c}   \}\\
    &= \inf  \{ c \in \R \, | \, \forall \,  \psi \in \mathcal{P}(\lambda,f^\eta), \, \exists \, \g \in \Gamma, \, \forall \w, \, f(\lambda(\w,\g))^\eta \geq  ({f(\psi(\w))^{\eta}})^c \},
\end{align*}
for $f = e^{-u}$, hence $\widetilde{c}(\eta)$ coincides with $c(\eta)$ defined in \eqref{eq-c-f-dep}.

We give equivalent forms of \eqref{eq-best-regret-bound-possible}. First, since $u^{-1}(x)$ is increasing, \eqref{eq-best-regret-bound-possible} is equivalent to
\begin{align*}
     u^{-1} \left(\sum_{t=1}^T \Tilde{\lambda}(\w_t,\gamma_t)) \right) \le u^{-1}\left( c \sum_{t=1}^T \Tilde{\lambda}(\w_t,\xi_t(\theta))) + a\ln\(|\Theta|\)\right).
\end{align*}
Furthermore, $\Tilde{\lambda} = u\circ \lambda$ and $f(x) = e^{-u(x)}$, so \eqref{eq-best-regret-bound-possible} is equivalent to
\begin{align*}
    \aggregation_T(\lambda(\omega_1, \gamma_1),\ldots, \lambda(\omega_T, \gamma_T)) \le u^{-1}( c \ u(\A_T(\theta^*)) + a  \ln(n) ).
\end{align*}
\end{proof}
\begin{remark}
    The optimality of the Aggregating Algorithm under quasi-sum aggregation only refers to this specific definition of global prediction game. Note, if $c > 1$ the tightness of the regret-like bound depends on the performance of the experts. Standard $O(\sqrt{T})$-regret algorithms in learning under expert advice, \eg, Exponential Weighting Algorithm, can potentially perform better, in terms of less loss, than the Aggregating Algorithm, even though the Aggregating Algorithm is optimal in the sense specified above \citep[\p 14]{cesa2006prediction}.
\end{remark}

\section{How aggregation changes prediction}
In the previous sections we have argued that the Aggregating Algorithm is generally applicable for losses interacting nicely with ``reasonable'' aggregation functions. However, it is still unclear how the aggregation influences the actual predictions made by the Aggregating Algorithm. This question brings us closer to understand why we should care about different general aggregation functionals.

The Aggregating Algorithm is, that is the finding of the previous sections, very generally applicable for losses nicely interacting with ``reasonable'' aggregation functionals. It is, however, still unclear how the aggregation influences the actual predictions made by the Aggregating Algorithm. %This question brings us closer to understand why we should care about different general aggregation functionals.

We qualitatively approach this question in three ways. First, we propose to interpret the generator functions of aggregations as utility functions. Then, we elaborate on the interpretation of weighting profiles associated with every general aggregation functional. Finally, we illustrate in an experiment that aggregations can express the forecaster's attitude towards losses. This analysis is not intended to provide a definite answer but rather to suggest potential paths for future research.
We approach this question in three ways. First, we propose to interpret the generator functions of aggregations as utility functions. Then, we elaborate on the interpretation of weighting profiles associated with every general aggregation functional. Finally, we illustrate in an experiment that aggregations can express the forecaster's attitude towards losses.

\subsection{Aggregation and utility of losses}
\label{Aggregation and Utility of Losses}
Aggregation functions for losses are, under mild conditions, quasi-sums $\Q^u$ (\cf Lemma~\ref{lemma-axiom-char}). On one hand, as we have shown, the $u$-quasi-sum of losses $\loss$ in the regret bound is tantamount to summing up distorted losses $u \circ \loss$ (Corollary~\ref{corollary:change of variables of the AA}). On the other hand, $u$ \emph{can be interpreted as} the negative utility function for the losses $\loss$ of the predictors. It expresses the dis-satisfaction of the learner to incur certain losses. Therefore, whether we talk about a certain choice of negative utility of losses or whether we talk about $u$-quasi-sums as aggregation functional does not make a difference.

\begin{figure}
    \centering
    \def\svgwidth{0.4\columnwidth}
    %% Creator: Inkscape 1.3.2 (091e20e, 2023-11-25), www.inkscape.org
%% PDF/EPS/PS + LaTeX output extension by Johan Engelen, 2010
%% Accompanies image file 'utitilities.pdf' (pdf, eps, ps)
%%
%% To include the image in your LaTeX document, write
%%   \input{<filename>.pdf_tex}
%%  instead of
%%   \includegraphics{<filename>.pdf}
%% To scale the image, write
%%   \def\svgwidth{<desired width>}
%%   \input{<filename>.pdf_tex}
%%  instead of
%%   \includegraphics[width=<desired width>]{<filename>.pdf}
%%
%% Images with a different path to the parent latex file can
%% be accessed with the `import' package (which may need to be
%% installed) using
%%   \usepackage{import}
%% in the preamble, and then including the image with
%%   \import{<path to file>}{<filename>.pdf_tex}
%% Alternatively, one can specify
%%   \graphicspath{{<path to file>/}}
%% 
%% For more information, please see info/svg-inkscape on CTAN:
%%   http://tug.ctan.org/tex-archive/info/svg-inkscape
%%
\begingroup%
  \makeatletter%
  \providecommand\color[2][]{%
    \errmessage{(Inkscape) Color is used for the text in Inkscape, but the package 'color.sty' is not loaded}%
    \renewcommand\color[2][]{}%
  }%
  \providecommand\transparent[1]{%
    \errmessage{(Inkscape) Transparency is used (non-zero) for the text in Inkscape, but the package 'transparent.sty' is not loaded}%
    \renewcommand\transparent[1]{}%
  }%
  \providecommand\rotatebox[2]{#2}%
  \newcommand*\fsize{\dimexpr\f@size pt\relax}%
  \newcommand*\lineheight[1]{\fontsize{\fsize}{#1\fsize}\selectfont}%
  \ifx\svgwidth\undefined%
    \setlength{\unitlength}{540bp}%
    \ifx\svgscale\undefined%
      \relax%
    \else%
      \setlength{\unitlength}{\unitlength * \real{\svgscale}}%
    \fi%
  \else%
    \setlength{\unitlength}{\svgwidth}%
  \fi%
  \global\let\svgwidth\undefined%
  \global\let\svgscale\undefined%
  \makeatother%
  \begin{picture}(1,0.99814818)%
    \lineheight{1}%
    \setlength\tabcolsep{0pt}%
    \put(0,0){\includegraphics[width=\unitlength,page=1]{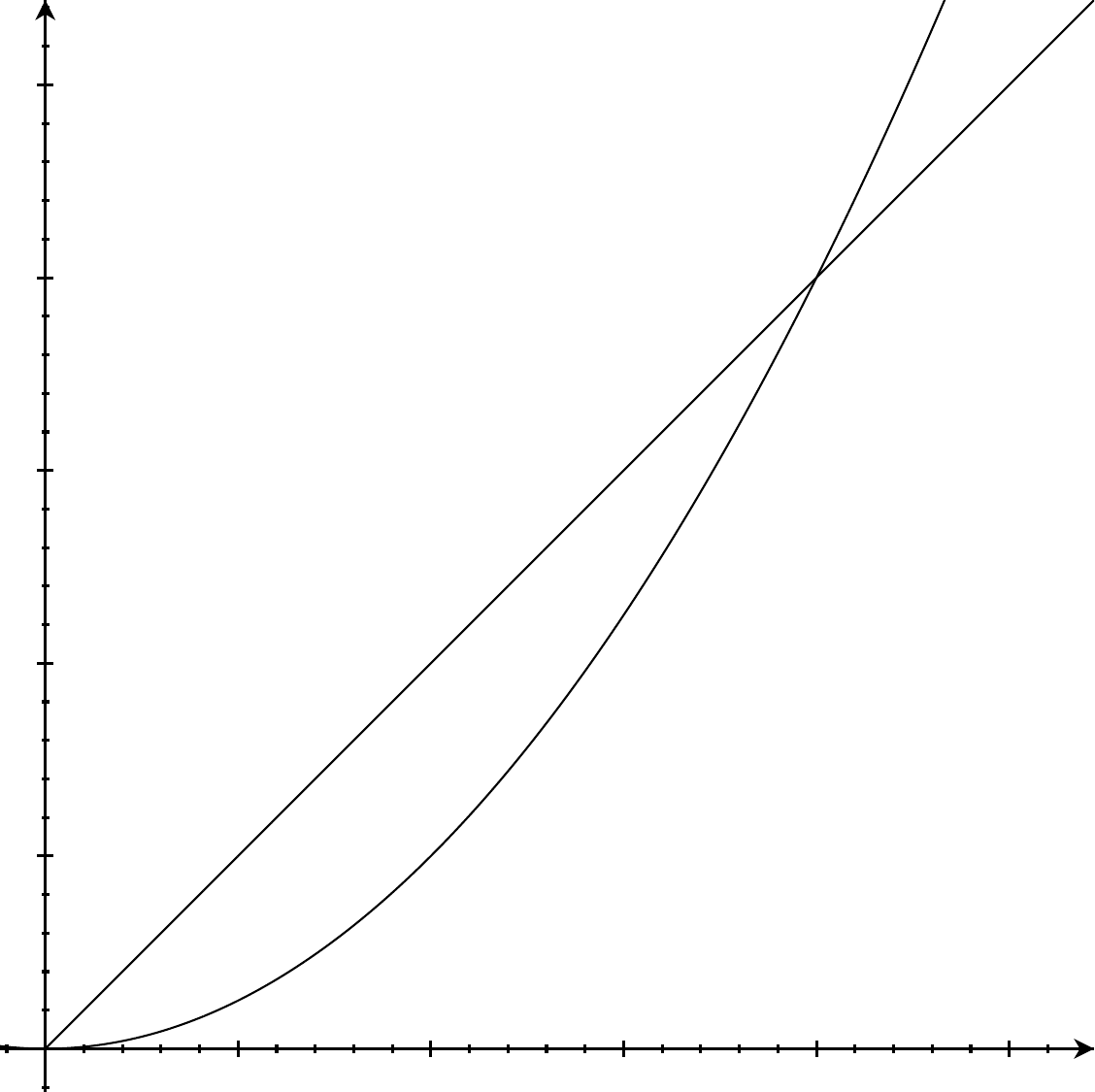}}%
    \put(0.7980913,0.72315587){\color[rgb]{0,0,0}\makebox(0,0)[lt]{\lineheight{1.25}\smash{\begin{tabular}[t]{l}$u(x) = x$\end{tabular}}}}%
    \put(0.91492199,0.06433902){\color[rgb]{0,0,0}\makebox(0,0)[lt]{\lineheight{1.25}\smash{\begin{tabular}[t]{l}$\lambda$\end{tabular}}}}%
    \put(0.06025054,0.94028739){\color[rgb]{0,0,0}\makebox(0,0)[lt]{\lineheight{1.25}\smash{\begin{tabular}[t]{l}$u(\lambda)$\end{tabular}}}}%
    \put(0.53228124,0.92650972){\color[rgb]{0,0,0}\makebox(0,0)[lt]{\lineheight{1.25}\smash{\begin{tabular}[t]{l}$u(x) = x^2$\end{tabular}}}}%
    \put(0,0){\includegraphics[width=\unitlength,page=2]{utitilities.pdf}}%
  \end{picture}%
\endgroup%

    \caption{Comparative Example of Linear and Squared Utility. { The horizontal axis denotes the loss value . The vertical axis the negative utility of the loss. We compare the negative utility function $u(x) = x$ to $u(x)=x^2$. In particular, for two values highlighted by a darkgreen arrow, low value, and an orange-brown arrow, high value.}}
    \label{fig:util comparison}
\end{figure}

Let us consider a comparative example: the simple negative utility function $u(x)=x$ corresponds to the standard sum. The negative utility function $u(x) = x^2$ generates the Euclidean norm aggregation. Compared to summation, in the Euclidean norm large loss values contribute relatively more to the result than small loss values. The analogous statement is true for the utility functions.
As a negative utility function $u(x) = x^2$, large loss values hurt, since higher negative utility (\cf orange-brown arrow in Figure~\ref{fig:util comparison}), relatively more than small loss values (\cf darkgreen arrow in Figure~\ref{fig:util comparison}).

More generally, it is true that risk-avoider prefer convex $u$, \ie, high losses are up-valued, low losses are down-valued. In contrast, risk-taker consider concave $u$, which means that low losses are up-valued and high losses are down-valued (\cf \citep{winkler1970nonlinear}, concavity and convexity are switched therein for reasons of sign flip). Hence, we can conclude: \textbf{the type of aggregation captures the attitude of the forecaster towards losses.}

\subsection{Weighting profiles describe the updating step}
Central to the Aggregating Algorithm is the weight updating. The weighting profile $f$ in Definition~\ref{def:weighting profile} determines how much the expert $\theta$ contributes to the next \pseudoprediction depending on the incurred loss in the current time step. To remind the reader, step (1) in the APA-QS (Algorithm~\ref{alg:APA-QS}),
\begin{align*}
    P_{t} (\theta) &= f(\lambda(\w_{t},\xi_{t}(\theta))) P_{t-1}(\theta),
\end{align*}
where under the common assumptions in this work, $f(x) = e^{-u(x)}$. Hence, the change of aggregation $\Q^u$ amounts to the change of weighting profile. 

\begin{table}
    \centering
    \caption{Aggregations, their corresponding utility function and weighting profiles for different learning rates (lightseagreen: $\eta = 0.001$, limegreen: $\eta = 0.5$, mediumseagreen: $\eta = 1$, seagreen: $\eta = 2$, darkgreen: $\eta = 10$, darkslategray: $\eta = 100$). The loss histogram shows the number of predictions (y-axis) incurring the loss (x-axis) in a certain bin. The blue line depicts the average loss. \\}
    \label{tab:aggregations - and weighting profiles}
    \begin{tabular}{cccc}
        \makecell[cc]{\textbf{Aggregation}\\ $u(x)=$} & Weighting Profile & Loss Histogram (Weather - Zugspitze) \\ \hline
        \makecell[cc]{\textbf{$L_{0.5}$-norm}\\ $\sqrt{x}$}  & \makecell[cc]{\includegraphics[width=100pt]{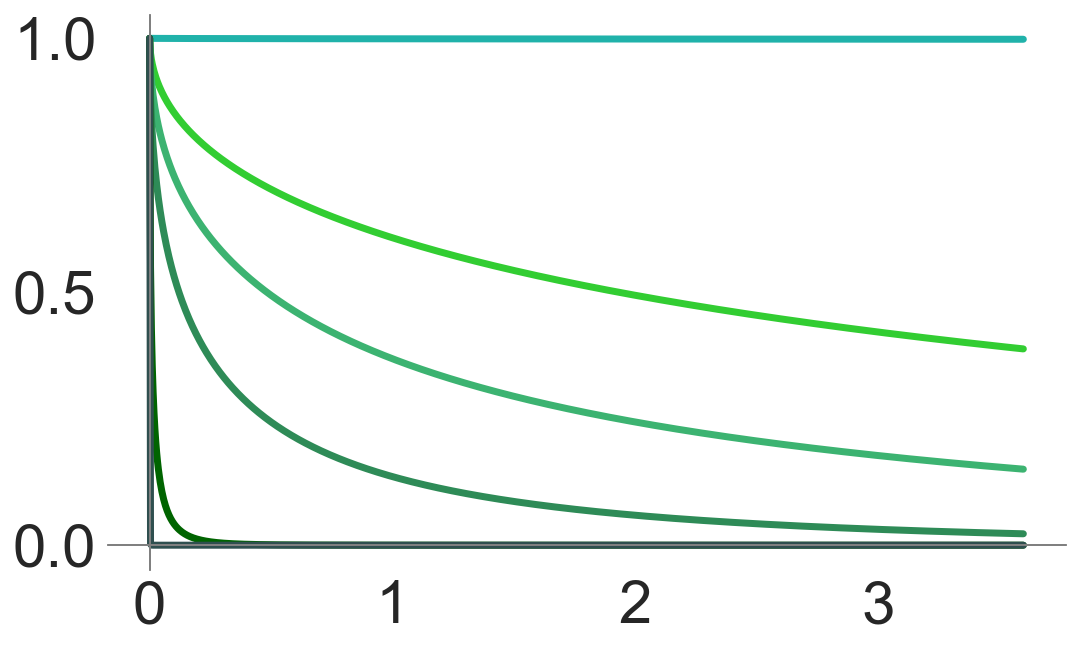}} & \makecell[cc]{\includegraphics[width=180pt]{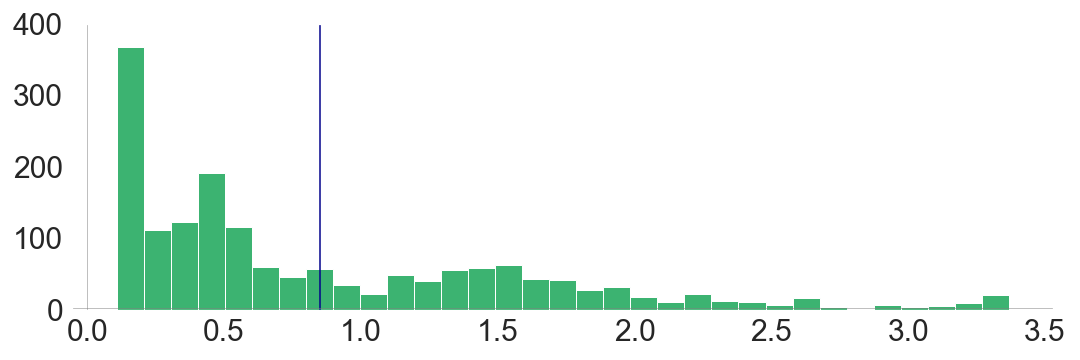}}\\ \hline
        \makecell[cc]{\textbf{Sum}\\$x$} & \makecell[cc]{\includegraphics[width=100pt]{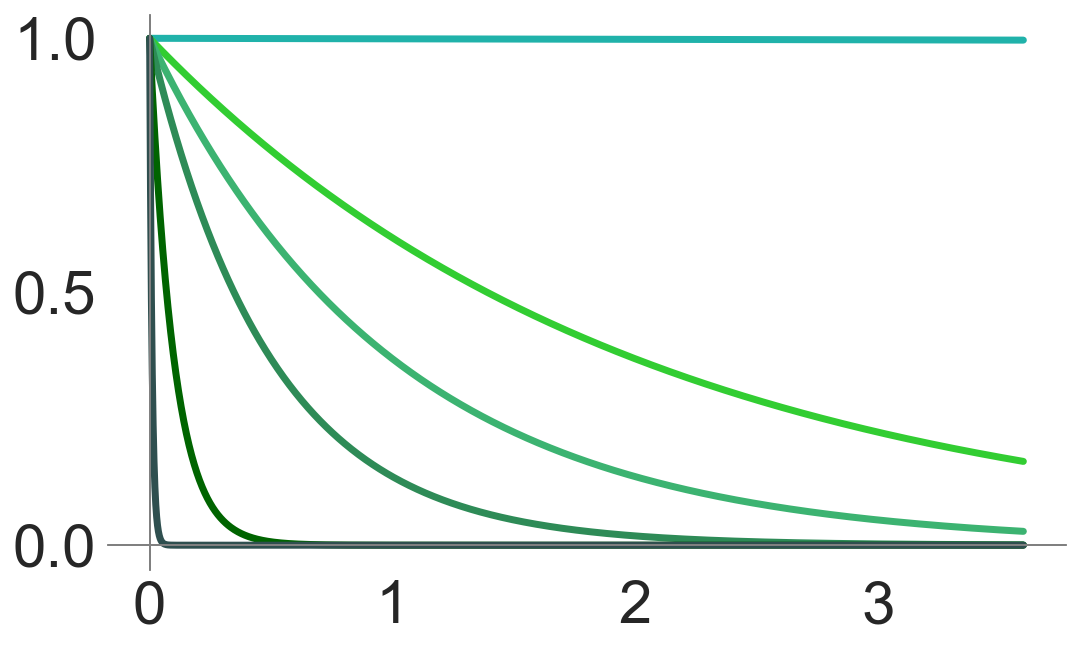}} & \makecell[cc]{\includegraphics[width=180pt]{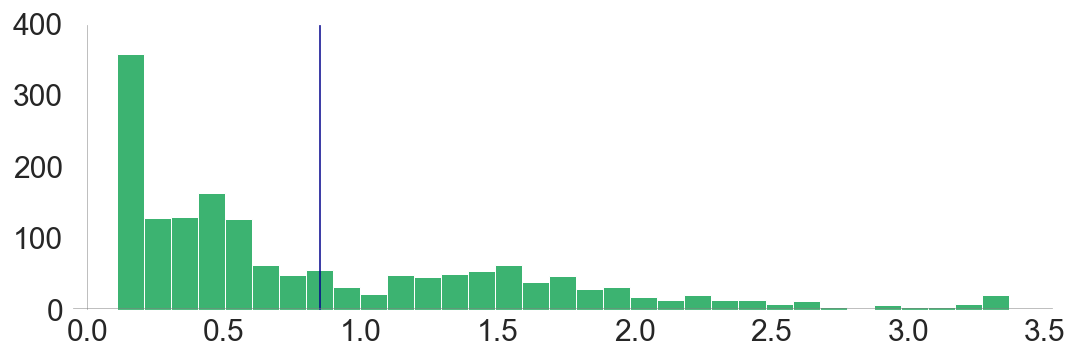}}\\ \hline
        \makecell[cc]{\textbf{$L_2$-norm} \\ $x^2$} & \makecell[cc]{\includegraphics[width=100pt]{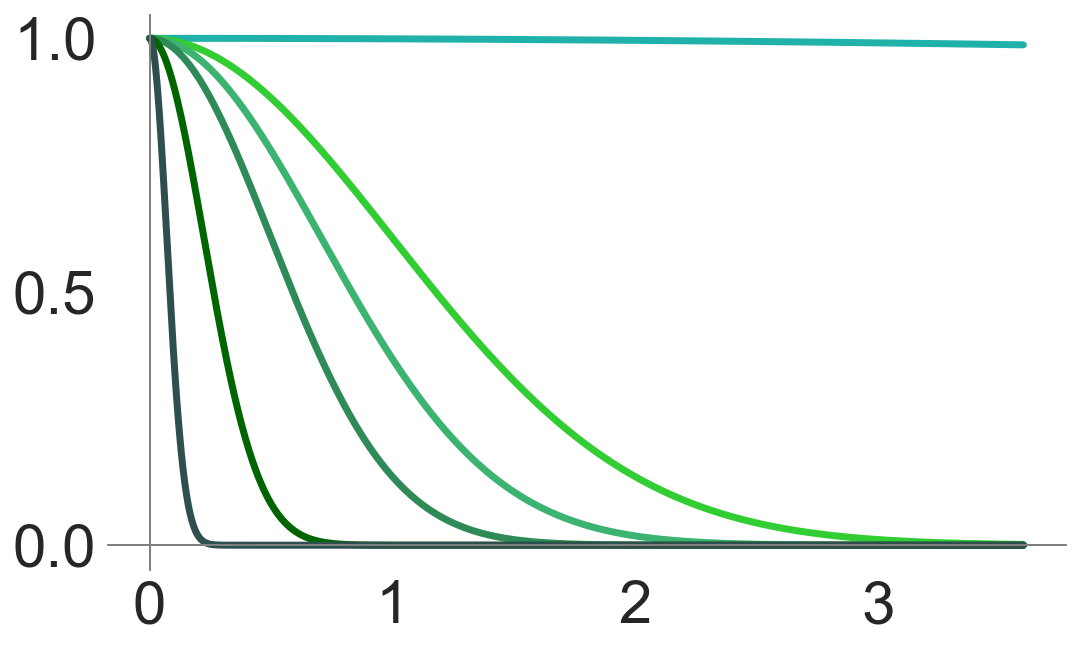}} & \makecell[cc]{\includegraphics[width=180pt]{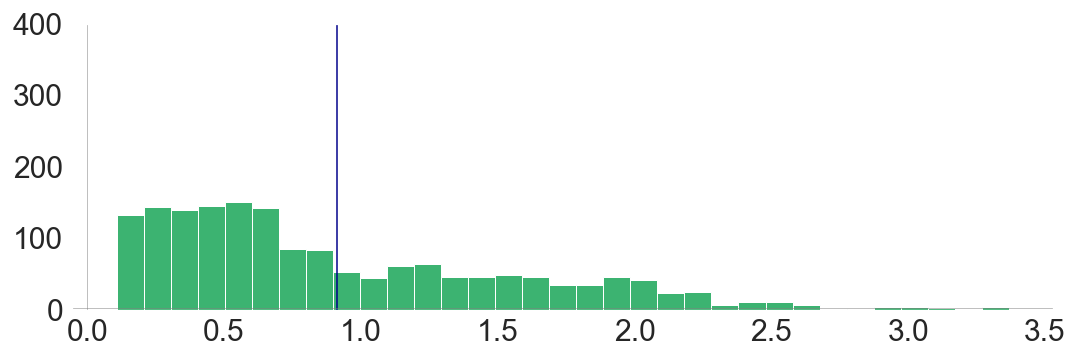}}\\ \hline
        \makecell[cc]{\textbf{$L_{10}$-norm} \\ $x^{10}$} & \makecell[cc]{\includegraphics[width=100pt]{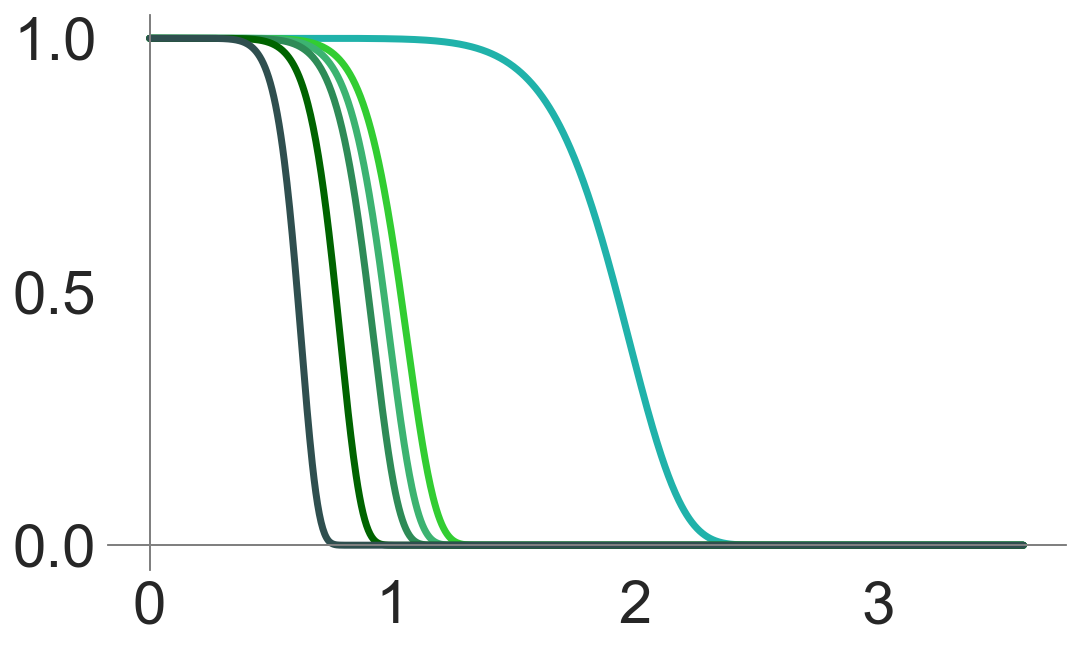}} & \makecell[cc]{\includegraphics[width=180pt]{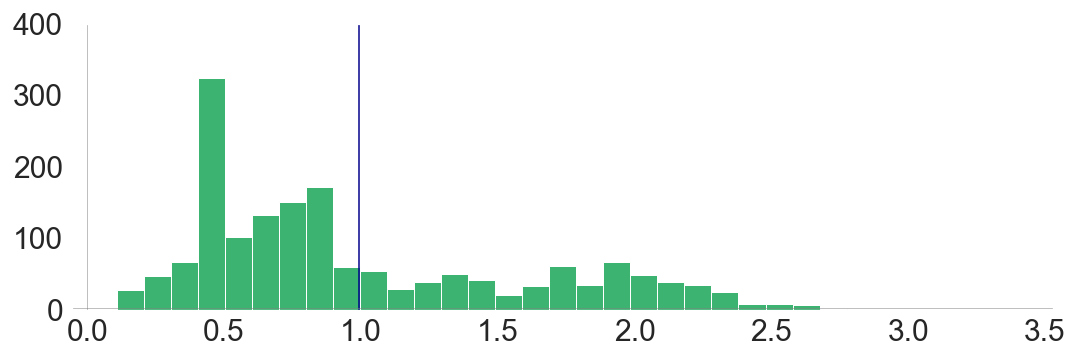}}\\ \hline
        \makecell[cc]{\textbf{Focal}\\ \textbf{aggregation}\\ $(1-\exp(-x))^2x$} & \makecell[cc]{\includegraphics[width=100pt]{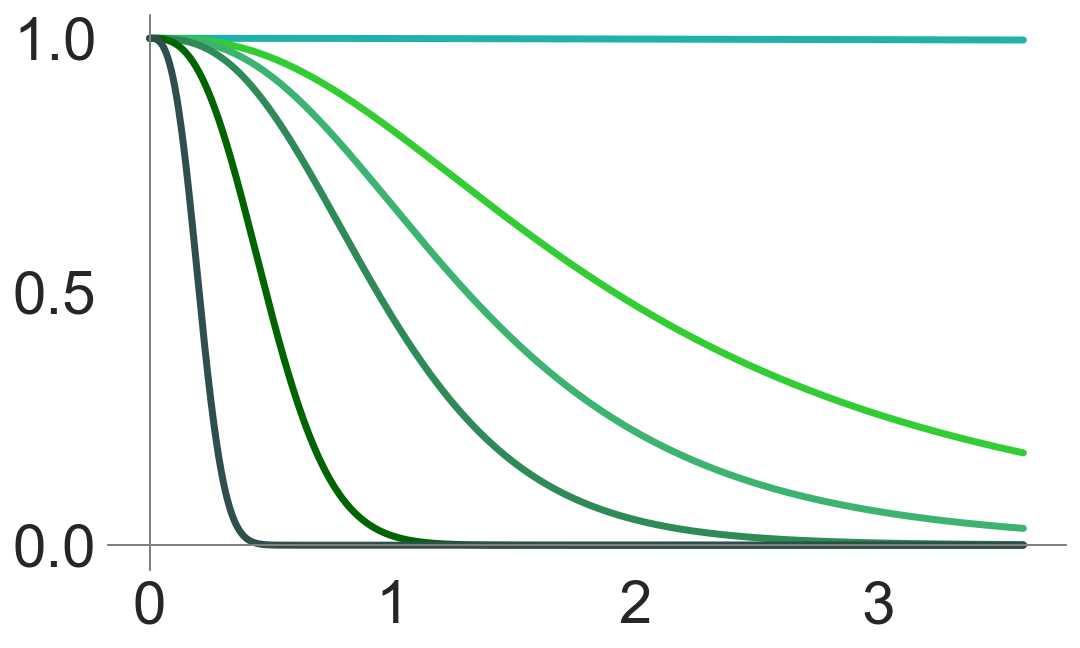}} & \makecell[cc]{\includegraphics[width=180pt]{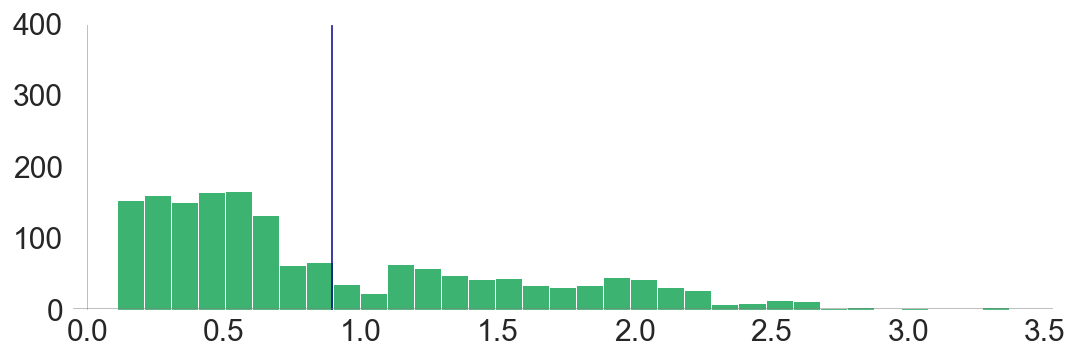}}\\
    \end{tabular}
\end{table}

For a fixed loss function we can analyze and compare weighting profiles for different aggregation functions. Table~\ref{tab:aggregations - and weighting profiles} provides a short list of potential aggregation functions, their corresponding utility (see Section~\ref{Aggregation and Utility of Losses}) and weighting profile. We use the term focal aggregation to emphasize that the composition of the corresponding $u$ with the log-loss recovers the often used focal loss (with $\gamma = 2$) \citep{lin2017focal}.
Observe that for $L_p$-norm aggregations switching the value of $p$ from less than $1$ to strictly bigger than $1$ drastically changes the shape of the weighting profile. Let us shortly compare the $L_2$-norm and sum. For the sake of simplicity, we focus on the arbitrary learning rate $\eta = 0.5$. The weighting profile corresponding to the $L_2$-norm punishes higher losses stronger than smaller losses compared to the sum. This, however, comes with the cost that for small losses the $L_2$ weighting profile does not finely distinguish between good and even better experts. Both of them get nearly updated with the same weight, in contrast to the sum.

Comparisons of weighting profiles require to fix a loss a priori. The domain and distribution of the loss values themselves strongly interact with the choice of aggregation. For instance, the Brier score only provides values between $0$ and $1$. Thus, the weighting profile beyond $1$ on the x-axis is irrelevant for the comparison. Hence, conclusive interpretations are only possible for fixed losses and different aggregations. Note that mixability might not be maintained for arbitrary combinations of losses and weighting profiles. For this reason, we go beyond this qualitative interpretation of the effect of aggregations and illustrate via an experiment how the aggregation changes the performance of the Aggregating Algorithm in Section~\ref{its not only theory}.

\subsection{Weather prediction via Aggregating Algorithm}
\label{its not only theory}
The preceding discussions suggest that aggregation, weighting profile and utility are essentially different facets of the same object. We earlier asked, how does the type of aggregation change the behavior of the Aggregating Algorithm? The proposed interpretations as weighting profile and utility lead us to the following three hypothesis, which we substantiate by a real-world data prediction experiment.
\begin{enumerate}[{\bf H-(a)}]
    \item Convex additive generators express the aversion towards extreme losses.
    \item Concave additive generators express the risk-loving behavior in terms of accepting extreme losses, but seeking for the ``perfect'' predictor.
    \item Focal aggregation expresses aversion towards extreme losses.
\end{enumerate}
We provide the log-loss profile of several applications of the AA-QS for different aggregations on a sequential weather classification task. We use the Aggregating Algorithm to aggregate probabilistic predictions of 9 simple classification algorithms. 

In more detail, we run a tabular classification task on weather data collections. The data is curated and provided by the DWD (German Weather Agency). The data collections are publicly available at \url{https://opendata.dwd.de/climate_environment/CDC/observations_germany/climate/daily/kl/historical/}. For the file names see Table~\ref{tab:weather data sets}. Note that the files are constantly updated, hence the file names potentially change. The file names are of the format \texttt{tageswerte\_KL\_[location identifier]\_[start date]]\_[end date]\_hist.zip}. Each collection provides daily measurements of weather-related attributes for one place. In every collection we deleted days with missing values. For statistic of the data collections see Table~\ref{tab:weather data sets}.

\begin{table}
    \centering
        \caption{Weather data collections from the DWD (German Weather Agency).\\}
    \label{tab:weather data sets}
    \begin{tabular}{cccc}
        Place & File\\ \hline
        Zugspitze & \texttt{tageswerte\_KL\_05792\_19000801\_20221231\_hist.zip}\\
        Potsdam & \texttt{tageswerte\_KL\_03987\_18930101\_20231231\_hist.zip}\\
        Putbus & \texttt{tageswerte\_KL\_04024\_18530701\_20231231\_hist.zip}\\
        Weissenburg-Emetzheim& \texttt{tageswerte\_KL\_05440\_18790101\_20231231\_hist.zip}\\
    \end{tabular}
    
    \begin{tabular}{cccc}
    \\
        Place & Days with Missing Values & Final Number of Days (Train/Test) \\ \hline
        Zugspitze & 15 & 8386 (6708/1678) \\
        Potsdam & 7 & 8759 (7007/1752) \\
        Putbus & 431 & 8323 (6658/1665) \\
        Weissenburg-Emetzheim & 22 & 8744 (6995/1749) 
    \end{tabular}
\end{table}

Based on daily average air pressure, average temperature, average relative humidity, maximum temperature, minimum temperature and date we train 9 classifiers to distinguish between four classes of weather: cloudy, rainy/snowy, sunny and unsettled, which are defined according to Table~\ref{tab:weather classes}.
\begin{table}
    \centering
        \caption{Definition of weather classes.\\}
    \label{tab:weather classes}
    \begin{tabular}{ccc}
         & Precipitation $\le 2mm$ & Precipitation $> 2mm$ \\ \hline
        Sunshine Hours $> 4h$ & sunny & unsettled \\
        Sunshine Hours $\le 4h$ & cloudy & rainy/snowy
    \end{tabular}

\end{table}
We use the first (in chronological order) $80 \%$ of the data points as training set for the following classifiers provided by the \texttt{scikit learn} package: logistic regression (LR), gaussian naive bayes (NB), support vector machine (SVC), linear model with stochastic gradient descent (SGD), decision tree (DT), $k$-nearest neighbors (KNN), random foreast (RF), bagging on decision trees (BAGGING), gradient boosting on decision trees (GB). %For details and code see \url{https://github.com/rabeUni/Aggregating-Algorithm-for-Quasi-Sums---Weather-Data-Collection}.

Then, we run the classifiers on the remaining $20\%$ of the data. We apply the Aggregating Algorithm for quasi-sums with the log-loss.\footnote{Note that the classifiers have not necessarily been trained using the log-loss function. This ``mismatch'' in classification tasks does not diminish the performance of the Aggregating Algorithm. However, it does allow for further optimization of the entire classification pipeline.} We use different aggregations in the Aggregating Algorithm as specified in Table~\ref{tab:aggregations - and weighting profiles}. We observed that the learning rate did not have a big impact on the loss histograms in our experiment. So slightly arbitrarily, we chose $\eta = 1$ for sum and focal aggregation, we chose $\eta = 2$ for $L_{0.5}$, $\eta = 0.001$ for $L_{10}$ and $\eta = 0.5$ for $L_2$. We remark that the chosen learning rates (or the aggregations) don't necessarily guarantee that $(f, \eta)$-mixability with respect to the aggregation holds for the log-loss function. However, Theorem~\ref{thm-GAPA-bound} still applies for all aggregations.

The code was run on MacBook Pro with Intel Core i9. However, the experiments only required a fraction of the compute power. We used \texttt{Python} 3.10.2, \texttt{scikit learn} 1.4.0, \texttt{pandas} 1.4.1, \texttt{nump}y 1.22.2, \texttt{matplotlib} 3.5.1.

For the loss histograms we used automatic binning on the interval $0.0$ to $3.4$. Higher values, which turned out to be \texttt{np.inf}, were cut off. For this reason, we introduced the $\infty$-bars in the plots where necessary. See Section~\ref{appendix:more experiments} for more experiments.

Table~\ref{tab:aggregations - and weighting profiles} summarizes our findings for the data collection from the Zugspitze. See Appendix~\ref{appendix:more experiments} for further experiments. The loss histograms particularly reveal the difference between convex and concave utility function. Concave utility functions (\ie, $L_{0.5}$ and Sum) lead to a high number of predictions with extremely small loss values, with the downside that some predictions incur a high loss. Convex utility functions (\ie, $L_{2}$ and Sum) damp the tails of high losses, \ie, only few predictions with high losses are made. On the other hand, there are many predictions made incurring small but suboptimal loss. In this realm, $L_{10}$ is the most extreme example in which many sub-optimal predictions are made, but the tails are banned. The focal aggregation leads to behavior largely similar to the one induced by $L_2$-aggregation. All three hypothesis about the change of prediction given a certain aggregation can be substantiated in this explorative study. An exhaustive experimental study would be required to provide more conclusive statements.

\section{Conclusion}
\label{conclusion}
In this paper, we put forward a general, axiomatical approach to loss aggregation in online learning. Analogous to the development in offline learning, but differently motivated, we show that a set of reasonable aggregations in online learning is characterized by the set of quasi-sums. It turns out that the AA can be adapted to deal with those general aggregations. Not only can we transfer the nice theoretical properties of the AA to its modified variant, we also provide experimental evidence that the choice of general aggregation determines the extreme-loss seeking or extreme-loss averse behavior of the AA. Hence, we span up a new dimension of choice, for which we think that the modified AA is just a starting point. We believe that similar modifications can be provided for other online learning algorithms such as weighted majority algorithm \citep{littlestone1994weighted}.

\subsection{Broader impact and position statement}
\label{ethical statement}
The generality of the learning under expert advice setting allows for the deployment of the adapted Aggregating Algorithm in a variety of settings, which do not exclude any malicious nor benevolent use. Note that the type of aggregation creates another choice parameter which, depending on the deployment context, might call for a participatory, democratic approach to the determination of the used quasi-sum. This contextualization requires further studies. We are convinced that our socialization has shaped our method and approach to research. We might have been ignorant to aspects of our work which against other socio-cultural backgrounds might miss or need reframing.

\section{Acknowledgements}
The authors appreciate and thank the International Max Planck Research School for Intelligent System (IMPRS-IS) for supporting Rabanus Derr. Part of this project was conducted when Rabanus Derr was visiting Aaron Roth at the University of Pennsylvania. The stay was supported by a fellowship of the German Academic Exchange Service (DAAD).
This work was funded in part by the Deutsche Forschungsgemeinschaft (DFG, German Research Foundation) under Germany’s Excellence Strategy — EXC number 2064/1 — Project number 390727645; it was also supported by the German Federal Ministry of Education and Research (BMBF): Tübingen AI Center.

%%%%%%%%%%%%%%%%%%%%%%%%%%%%%%%%%%%%%%%%%%%%%%%%%%%%%%%%%%%%

\bibliographystyle{plainnat}
\bibliography{main}

%%%%%%%%%%%%%%%%%%%%%%%%%%%%%%%%%%%%%%%%%%%%%%%%%%%%%%%%%%%%

\newpage
\appendix
\section{More Experiments}
\label{appendix:more experiments}
The following tables provide more examples of the same experiment as described in Section~\ref{its not only theory}. Note that in some of the experiments we observed $\infty$-losses due to numerical instabilities close to $0$.
\begin{table}[ht]
    \centering
    \caption{Aggregations, their corresponding utility function and loss histograms for the AA-QS on the weather data collection from Potsdam. The blue line depicts the average loss excluding $\infty$-losses.\\}
    \label{tab:aggregations - and weighting profiles - Potsdam}
    \begin{tabular}{ccc}
        \makecell[cc]{\textbf{Aggregation}\\ $u(x)=$} & Loss Histogram (Weather -- Potsdam) \\ \hline
        \makecell[cc]{\textbf{$L_{0.5}$-norm}\\ $\sqrt{x}$}  & \makecell[cc]{\includegraphics[width=220pt]{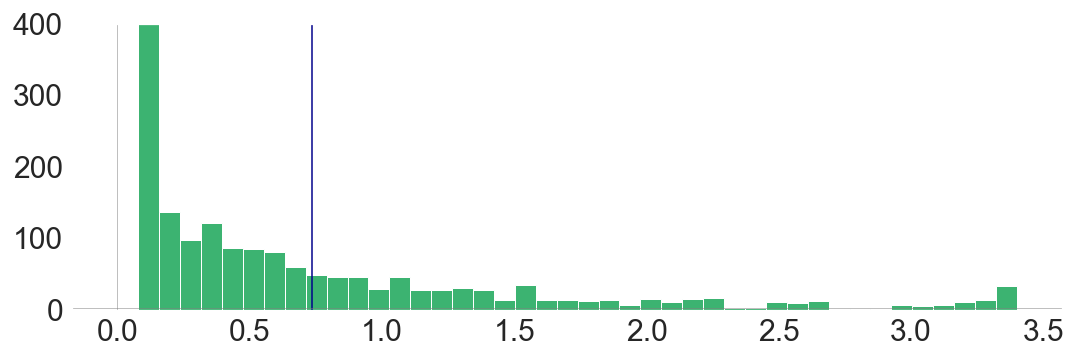}} \\ \hline
        \makecell[cc]{\textbf{Sum}\\$x$} & \makecell[cc]{\includegraphics[width=220pt]{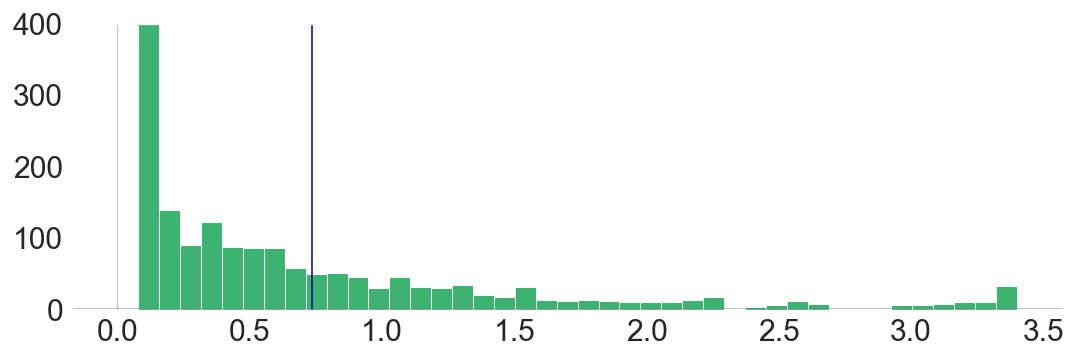}}\\ \hline
        \makecell[cc]{\textbf{$L_2$-norm} \\ $x^2$} & \makecell[cc]{\includegraphics[width=220pt]{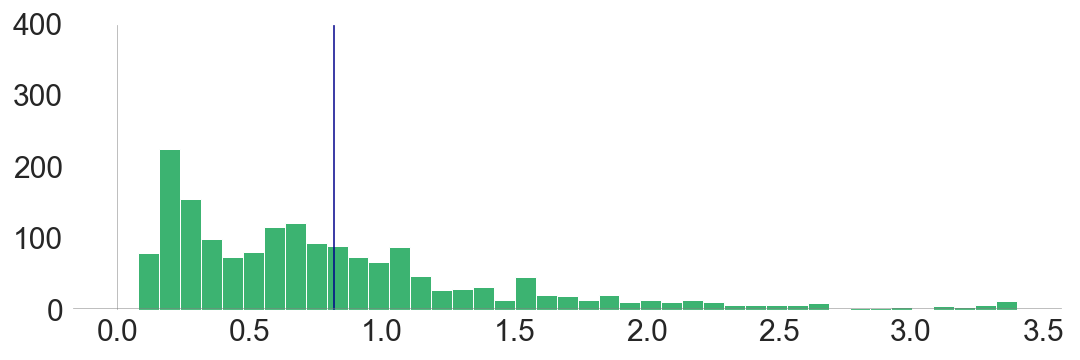}}\\ \hline
        \makecell[cc]{\textbf{$L_{10}$-norm} \\ $x^{10}$} & \makecell[cc]{\includegraphics[width=220pt]{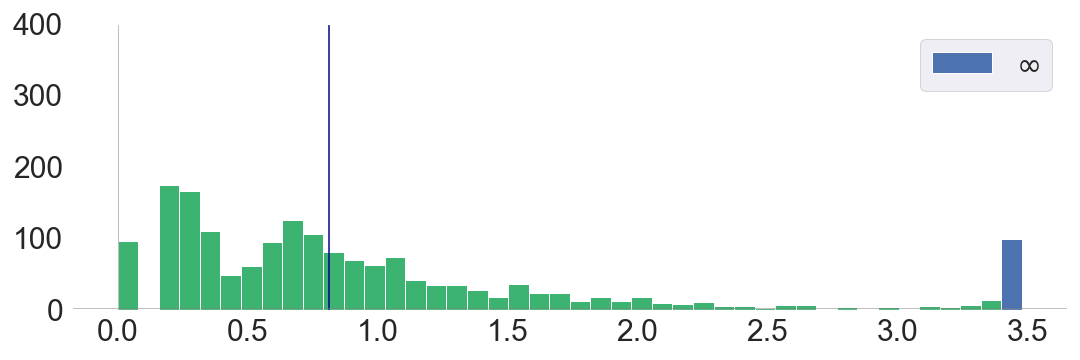}}\\ \hline
        \makecell[cc]{\textbf{Focal}\\ \textbf{aggregation}\\ $(1-\exp(-x))^2x$} & \makecell[cc]{\includegraphics[width=220pt]{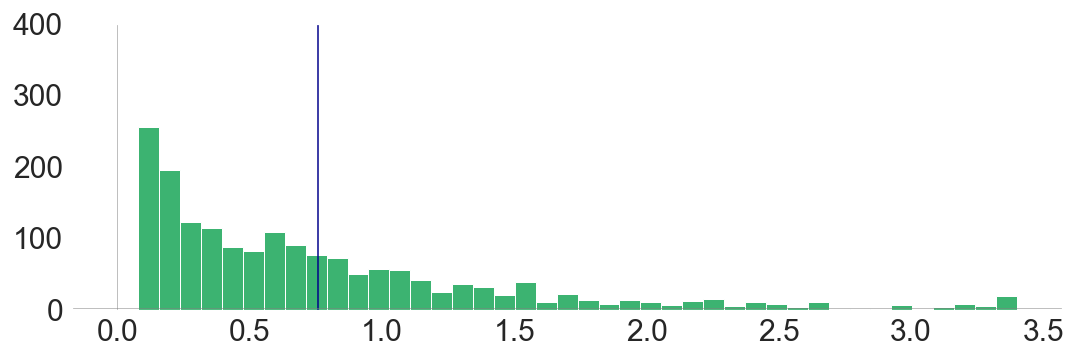}}\\
    \end{tabular}
\end{table}
\newpage
\begin{table}[ht]
    \centering
    \caption{Aggregations, their corresponding utility function and loss histograms for the AA-QS on the weather data collection from Putbus. The blue line depicts the average loss excluding $\infty$-losses.\\}
    \label{tab:aggregations - and weighting profiles - Putbus}
    \begin{tabular}{ccc}
        \makecell[cc]{\textbf{Aggregation}\\ $u(x)=$} & Loss Histogram (Weather -- Putbus) \\ \hline
        \makecell[cc]{\textbf{$L_{0.5}$-norm}\\ $\sqrt{x}$}  & \makecell[cc]{\includegraphics[width=260pt]{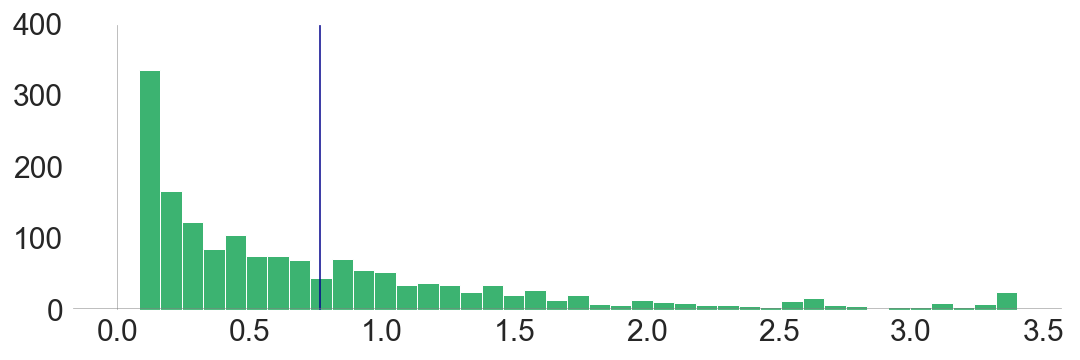}}\\ \hline
        \makecell[cc]{\textbf{Sum}\\$x$} & \makecell[cc]{\includegraphics[width=260pt]{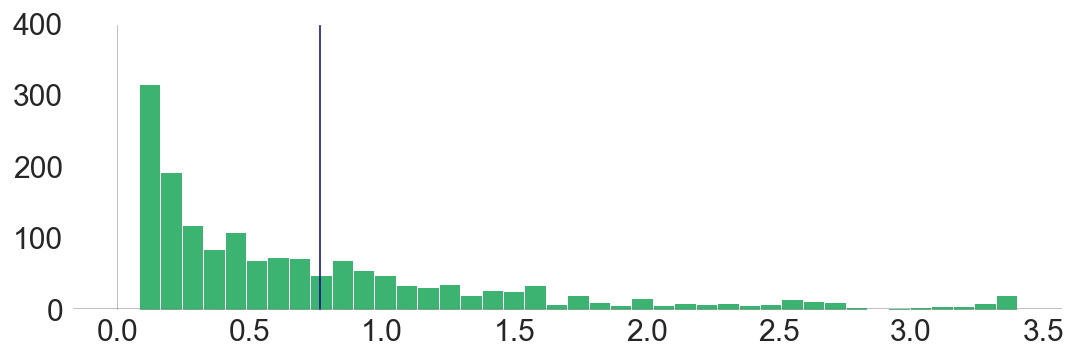}}\\ \hline
        \makecell[cc]{\textbf{$L_2$-norm} \\ $x^2$} & \makecell[cc]{\includegraphics[width=260pt]{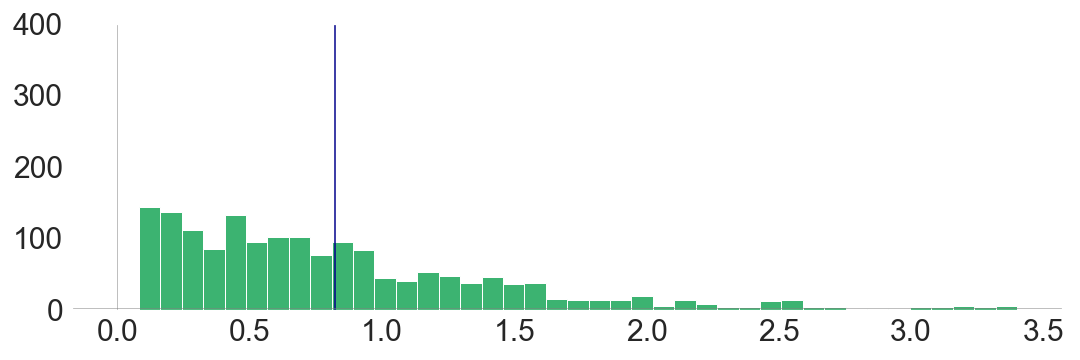}}\\ \hline
        \makecell[cc]{\textbf{$L_{10}$-norm} \\ $x^{10}$} & \makecell[cc]{\includegraphics[width=260pt]{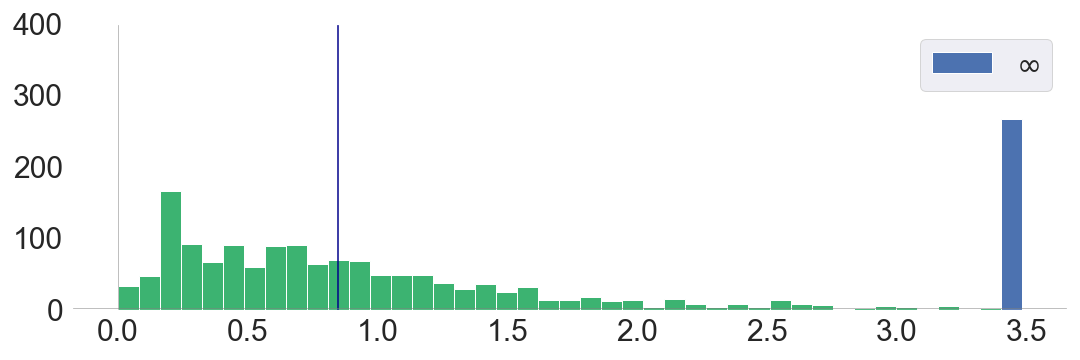}}\\ \hline
        \makecell[cc]{\textbf{Focal}\\ \textbf{aggregation}\\ $(1-\exp(-x))^2x$} & \makecell[cc]{\includegraphics[width=260pt]{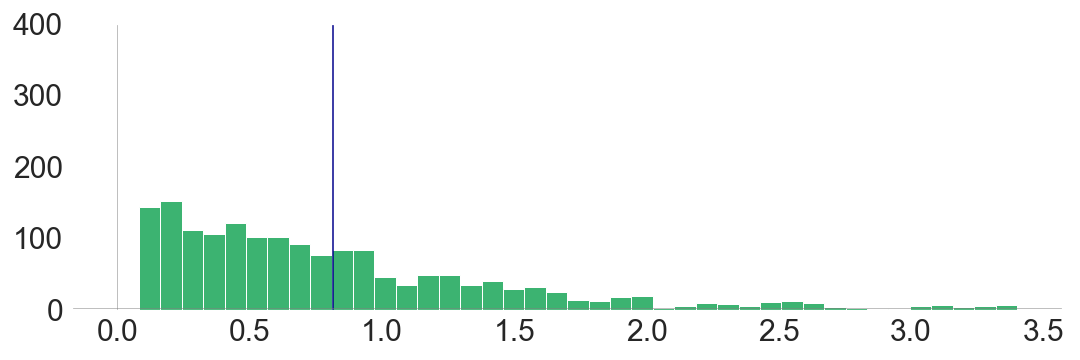}}\\
    \end{tabular}
\end{table}
\newpage
\begin{table}[ht]
    \centering
    \caption{Aggregations, their corresponding utility function and loss histogram for the AA-QS on the weather data collection from Weissenburg-Emetzheim. The blue line depicts the average loss excluding $\infty$-losses.\\}
    \label{tab:aggregations - and weighting profiles - weissenbrug-emetzheim}
    \begin{tabular}{ccc}
        \makecell[cc]{\textbf{Aggregation}\\ $u(x)=$} & Loss Histogram (Weather -- Weissenburg-Emetzheim) \\ \hline
        \makecell[cc]{\textbf{$L_{0.5}$-norm}\\ $\sqrt{x}$}  & \makecell[cc]{\includegraphics[width=260pt]{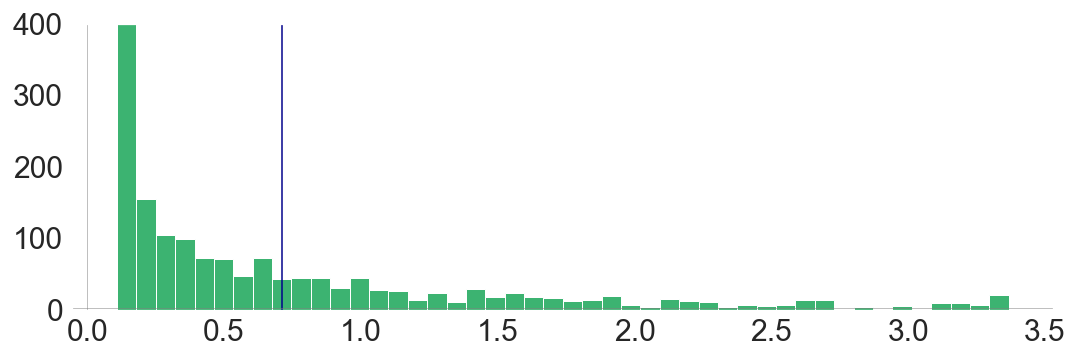}}\\ \hline
        \makecell[cc]{\textbf{Sum}\\$x$} & \makecell[cc]{\includegraphics[width=260pt]{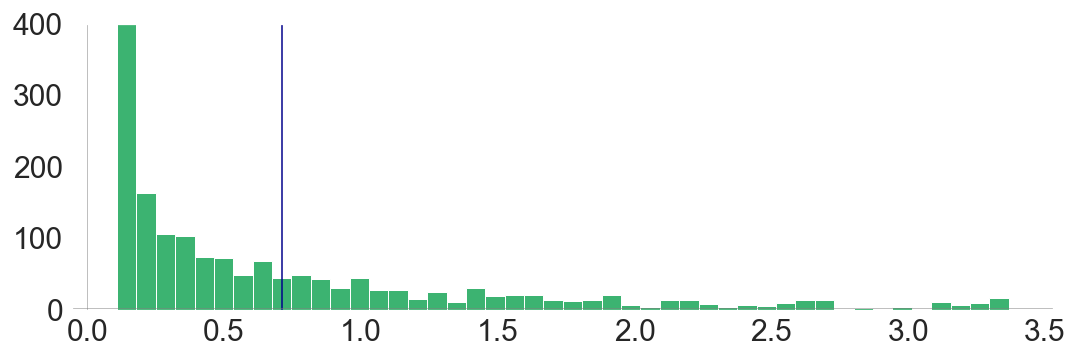}}\\ \hline
        \makecell[cc]{\textbf{$L_2$-norm} \\ $x^2$} & \makecell[cc]{\includegraphics[width=260pt]{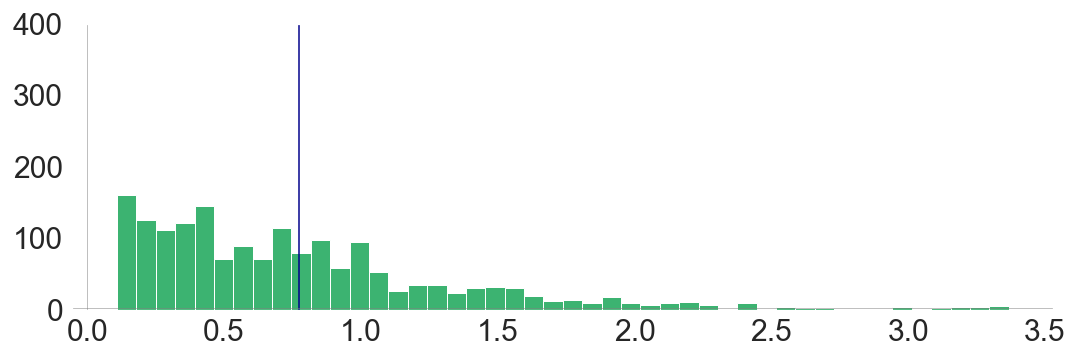}}\\ \hline
        \makecell[cc]{\textbf{$L_{10}$-norm} \\ $x^{10}$} & \makecell[cc]{\includegraphics[width=260pt]{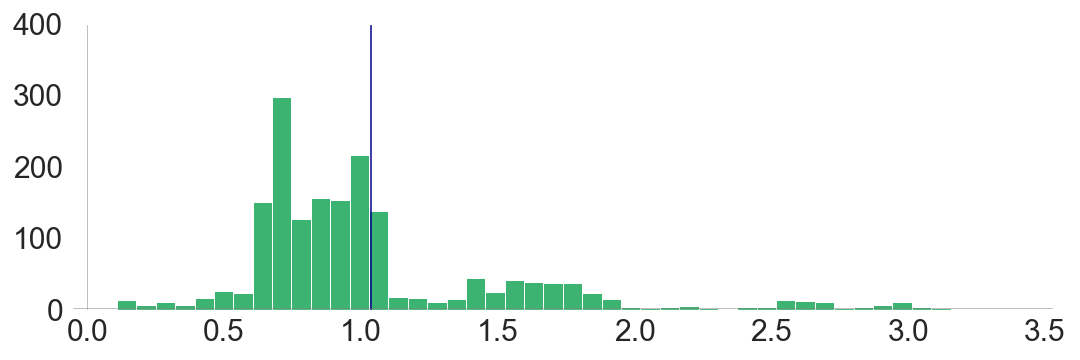}}\\ \hline
        \makecell[cc]{\textbf{Focal}\\ \textbf{aggregation}\\ $(1-\exp(-x))^2x$} & \makecell[cc]{\includegraphics[width=260pt]{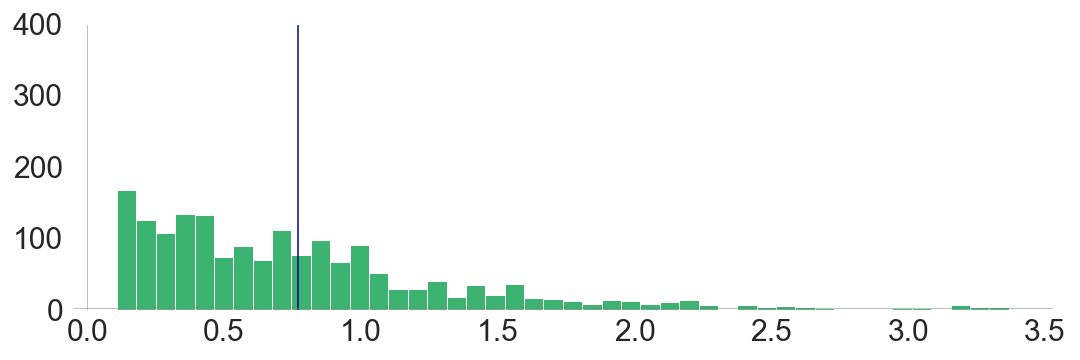}}\\
    \end{tabular}
\end{table}

\end{document}